\documentclass{article}

\usepackage{PRIMEarxiv}

\usepackage[utf8]{inputenc} % allow utf-8 input
\usepackage[T1]{fontenc}    % use 8-bit T1 fonts
\usepackage{hyperref}       % hyperlinks
\usepackage{url}            % simple URL typesetting
\usepackage{booktabs}       % professional-quality tables
\usepackage{amsfonts}       % blackboard math symbols
\usepackage{nicefrac}       % compact symbols for 1/2, etc.
\usepackage{microtype}      % microtypography
\usepackage{lipsum}
\usepackage{fancyhdr}       % header
\usepackage{graphicx}       % graphics
\graphicspath{{media/}}     % organize your images and other figures under media/ folder

\usepackage{amsfonts}
\usepackage{amsmath}
\usepackage{natbib}
\usepackage{xcolor}
\usepackage{multirow}
\usepackage{float}
\usepackage{algorithm}
\usepackage{algpseudocode}

\newcommand{\FDC}{\mathrm{FDC}}
\newcommand{\HD}{\mathrm{HD}}
\newcommand{\OS}{\mathrm{OrdSens}}
\title{JEPA as a Transferable Optimizer: Learning and Understanding Black-Box Optimization from Reward-Agnostic Pretraining
}

\author{
  Mayukh Debnath \\
  Robotics and Autonomous Systems (AI)\\
  Arizona State University\\
  Tempe, USA\\
  \texttt{mdebnath@asu.edu} \\
   \And
  Manan Roy Choudhury \\
  Department of Artificial Intelligence\\
  Arizona State University\\
  Tempe, USA\\
  \texttt{mroycho1@asu.edu} \\
}

\begin{document}
\maketitle

\begin{abstract}
Black-box optimization is ubiquitous: a control policy is scored by a rollout, a design by a simulation, a sequence by an assay. The absence of gradients has motivated a large family of hand-designed search rules, and more recently a family of learned ones, including methods that place a large language model directly in the optimization loop. Almost all of this work is evaluated on a single question, namely whether the learned optimizer outperforms its baselines. We ask a second question: what does a learned optimizer actually use? We introduce JEPA-PS (JEPA Policy Search), a black-box optimizer built from a Joint-Embedding Predictive Architecture that is pretrained on random rollouts using only parameter vectors and their scalar returns. Unlike prior work that meta-learns an update rule against a reward-prediction or gradient-matching objective, JEPA-PS is trained purely to predict the latent representation of the next point in an optimization trajectory, which separates \emph{what was tried} from \emph{how well it worked} and allows us to measure which of the two the optimizer depends on. At test time it reads a short evaluation history and proposes the next candidate in a single forward pass, with no task descriptions, no reward shaping, and no test-time gradients. We evaluate JEPA-PS on 15 control environments against 12 optimizers spanning classical evolutionary search, Bayesian optimization, restart-based CMA-ES, and an LLM optimizer. It is competitive with tuned classical baselines, outperforms the LLM optimizer by more than $2\times$ on Swimmer at $71\times$ lower per-step cost, and transfers zero-shot to environments held out of pretraining, and its latent representation is shown to carry information beyond simple history statistics via a controlled null-model comparison and a decoder-training ablation in which every trained decoder variant underperforms the default frozen one. Ablating the trained model then yields our main finding: the amount of history the optimizer requires is not a constant of the architecture but is elicited by the task. On smooth landscapes the full trajectory is used, while on rugged ones the best-seen point carries nearly all of the signal. This memory requirement correlates with the fitness-distance correlation (FDC) of the landscape ($r{=}0.112$, $n{=}7$), and an independent probe based on reward-ordering sensitivity does not corroborate this ($r{=}-0.024$, $n{=}7$). Code is available at \url{https://github.com/mdebnath22/jepa-props}.
\end{abstract}

% keywords can be removed

\section{Introduction}

Optimization is used everywhere, often without a clear view of what drives its success: a robot's gait is evaluated through a rollout, a protein through an assay, and a hyperparameter configuration through a training run \citep{snoek2012practical,mania2018simple}. Every black-box optimizer implicitly answers a question that is rarely asked: \emph{how much of its search history is needed to choose the next query?} Hand-designed methods answer it by construction. CEM refits a Gaussian to the elite set and discards the rest of the trajectory \citep{de2005tutorial}, CMA-ES compresses history into a covariance matrix and evolution paths \citep{hansen2016cma}, and Bayesian optimization retains every observation in a surrogate model \citep{snoek2012practical,shahriari2015taking}. Large-scale benchmarks find that no single choice dominates \citep{nevergrad}, which suggests that the right amount of memory depends on the problem, yet in these methods the designer, not the problem, fixes it.

Learned optimizers \citep{marcin2016learning,li2016learning,chen2017learning,metz2022velo,amos2023tutorial}, including LLM-based optimizers that read search histories as text \citep{yang2024large,song2025laser,qiu2024robomorph,lange2024large,zhou2026prompted,liu2024llambo,meindl2026gptopt}, are usually evaluated by whether they beat their baselines. That establishes effectiveness but not \emph{how} the history is used to pick the next candidate. The distinction matters most for in-context optimizers, which read a history and generate a candidate with no prescribed rule for using it. Their decisions let us ask which parts of the history they rely on and whether that reliance varies across problems.

We study this with \textbf{JEPA-PS} (JEPA Policy Search), a small in-context optimizer built on a joint-embedding predictive architecture \citep{lecun2022path,assran2023self}. It is pretrained on parameter vectors and scalar returns from random rollouts with a loss that predicts the \emph{latent representation} of the next trajectory point rather than the reward. At test time it reads a short history and proposes the next candidate in a single forward pass. Because its only input is that history, we can subset, reorder, permute, or corrupt it and observe how the proposal changes. JEPA-PS is therefore used here as an \emph{instrument} for measuring memory use, not as a new optimizer to adopt.

\paragraph{What we find.}
\begin{enumerate}
  \item \textbf{History use depends on the problem.} On CartPole the full trajectory beats the best-seen point by $22\%$ (mean best reward $321.4 \to 410.9$) and permuting reward order costs $37\%$; on MountainCar neither ablation has a measurable effect. We decompose this into four separable dependencies: the trajectory beyond its best point, parameter-space geometry, reward ordering rather than magnitude, and coarse temporal direction. Different problems switch each on or off to different degrees (Section~\ref{sec:analysis}).
  \item \textbf{The pattern is associated with landscape ruggedness.} History dependence is associated with fitness--distance correlation \citep{jones1995fitness} at $0.738$ across seven environments, and an independent reward-ordering probe at $0.832$ across five. These are small-sample associations, so we pair them with an explicit sensitivity analysis and treat them as preliminary (Section~\ref{sec:landscape}).
  \item \textbf{Transfer is real but limited.} A pre-registered, six-part investigation finds genuine, well-powered transfer across unseen CartPole physics variants, a mechanism probe that does not corroborate that context is the source, and no practical advantage over classical search started cold (Section~\ref{sec:transfer}).
  \item \textbf{JEPA-PS beats LLM optimizers, not classical search.} It is consistently cheaper per proposal than LLM optimizers (about $77\times$ on Swimmer, where it also scores more than twice as high) and usually competitive or better (Section~\ref{sec:results}). In a further comparison it beats ProPS and ProPS-Redundant while trailing plain CEM, and a restart-wrapped LLM result near the environment ceiling is attributable chiefly to the restart mechanism rather than reasoning (Section~\ref{sec:llm}).
  \item \textbf{No advantage over CMA-ES anywhere we tested.} A six-part battery finds a narrow representational signal but no case in which JEPA-PS's optimization performance exceeds a properly run CMA-ES (Appendix~\ref{app:cmaes_battery}).
\end{enumerate}

\paragraph{What we do not claim.}
JEPA-PS is not a general-purpose optimizer. It does not consistently outperform restart CMA-ES, run faster per step than classical rules, or transfer across structurally different environments. The transfer we document is not shown to be practically useful, since in a matched-budget comparison JEPA-PS performs worse than cold-started classical search. We also do not propose an optimizer that exploits the fitness--distance/memory association; the association itself is the contribution.

\paragraph{Contributions.}
\begin{enumerate}
  \item We develop a controlled ablation for measuring how much of its history a learned optimizer uses, and show that with the architecture fixed, history dependence varies across problems and reflects reward ordering, parameter geometry, broad temporal direction, and the trajectory after its best point.
  \item Across two mechanistically distinct interventions, we find an association between history dependence and fitness--distance correlation, accounting explicitly for sample size, leverage, and potential confounds.
  \item Ablations of the decoder and comparisons with latent-space CEM show that pretraining mainly provides reusable geometry rather than a proposal mechanism.
\end{enumerate}

% ═══════════════════════════════════════════════
\section{Related Work}
% ═══════════════════════════════════════════════
\paragraph{Learned and LLM-based optimizers.} One line of work trains networks directly as gradient-free or RL optimizers \citep{marcin2016learning,li2016learning,chen2017learning}; \citep{metz2022velo} scale this across thousands of tasks, and \citep{amos2023tutorial} frames the area as amortized optimization. A second line prompts an LLM with a running list of solutions and scores to optimize prompts, designs, programs, and policies \citep{yang2024large,song2025laser,qiu2024robomorph,meyerson2024language,romera2024mathematical,lange2024large,zhou2026prompted}. \citep{liu2024llambo} frame this explicitly as Bayesian optimization, using the LLM for candidate sampling \emph{and} surrogate ranking; we adapt this two-role structure as a baseline (Appendix~\ref{app:llambo}). \citep{meindl2026gptopt} instead fine-tune an open-weight LLM on BO trajectories directly -- a closer architectural relative of JEPA-PS than pure prompting, which we do not reproduce. \citep{ferreira2026centaur} independently find classical HPO beats pure-LLM optimizers on an LM-tuning task, consistent with our own comparisons (Section~\ref{sec:llm}). \citep{zhang2025pabbo} amortize black-box optimization across tasks, and \citep{alansari2026sparrow}'s rank-based, ordinal archives parallel our finding that JEPA-PS is invariant to monotone reward transformations.

\paragraph{Joint-embedding predictive architectures.} JEPA \citep{lecun2022path} predicts a target view's embedding from a context view rather than reconstructing inputs; I-JEPA, V-JEPA, and related latent-prediction objectives instantiate it across modalities \citep{assran2023self,bardes2024revisiting,baevski2022data2vec,grill2020bootstrap}, as do latent world models for control \citep{ha2018recurrent,hafner2023mastering}. To our knowledge, JEPA-PS is the first use of a JEPA-style latent-prediction objective to propose optimization parameters directly, rather than to represent states, images, or trajectories for downstream control.

\paragraph{Landscape analysis, population search, and restarts.} Fitness--distance correlation \citep{jones1995fitness} and random-walk autocorrelation \citep{weinberger1990correlated} characterize search difficulty from cheap random samples; we use them as optimizer-independent descriptors and ask whether they predict a learned optimizer's history use, a connection we are not aware of having been drawn. We adopt a standard island/swarm construction \citep{kennedy1995particle,whitley1999island,tomassini2005spatially} as an off-the-shelf stabilizer, not as a contribution. Domain randomization across physics variants of one environment \citep{tobin2017domain,peng2018sim} is structurally what our physics-variant transfer test (Section~\ref{sec:transfer}) does. Restarting on stagnation and running parallel instances, standard devices most notably in restart CMA-ES \citep{auger2005restart}, are inference-time interventions we apply to JEPA-PS, explicitly testing rather than assuming whether their benefit is attributable to pretraining or to the restart mechanism alone (Section~\ref{sec:transfer}; Appendix~\ref{app:cmaes_battery}).

% ═══════════════════════════════════════════════
\section{JEPA-PS as an Instrument}
\label{sec:method}
\begin{figure}[t]
\centering\includegraphics[width=0.9\linewidth,height=98cm,keepaspectratio]{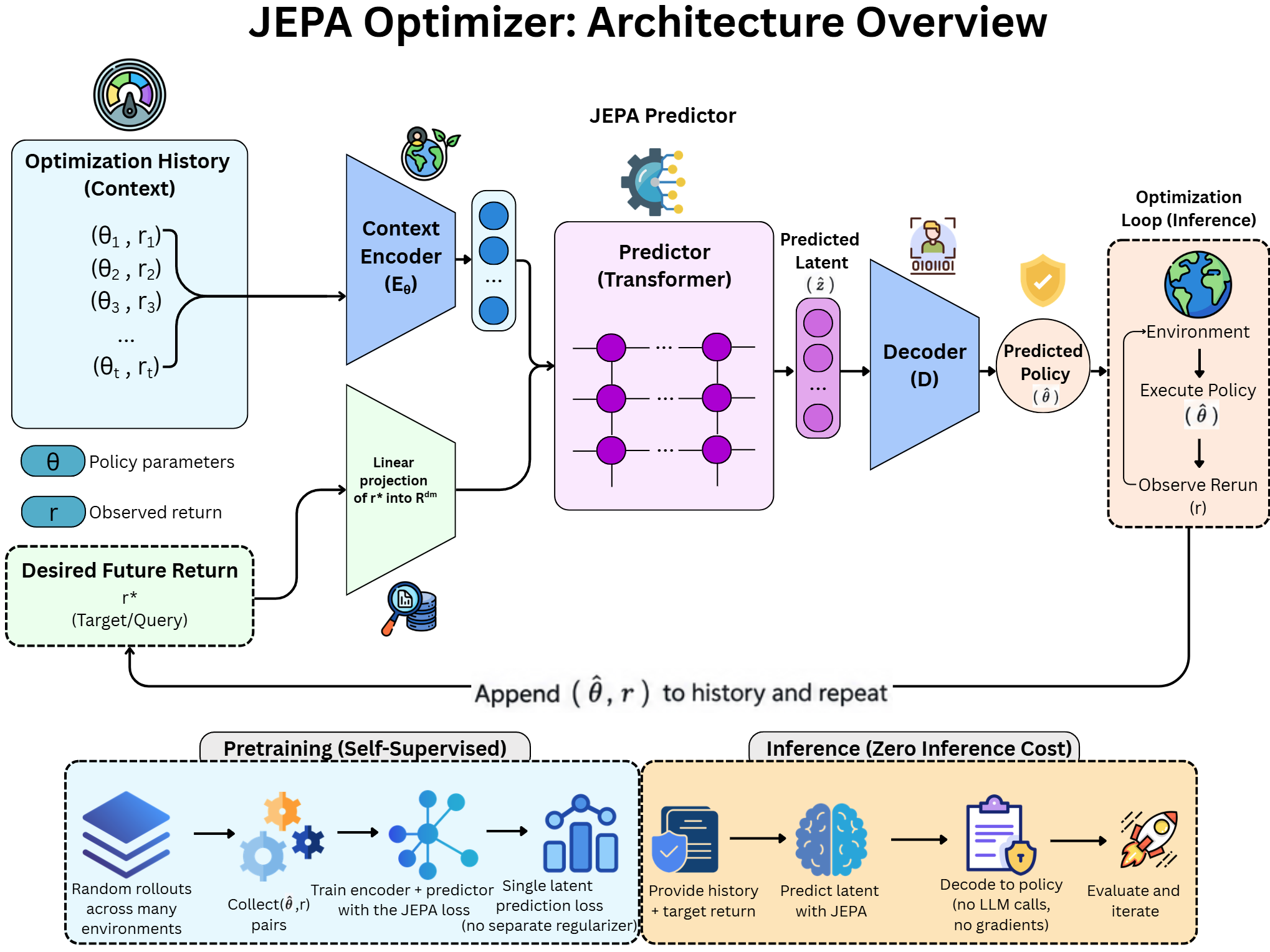}
\caption{Overview of the JEPA optimizer.}
\label{fig:arch}
\end{figure}

% ═══════════════════════════════════════════════
\paragraph{Setup.} We optimize $\theta\in\mathbb{R}^d$ of a policy $\pi_\theta$, $\theta^\ast=\arg\max_\theta J(\theta)$, $J(\theta)=\mathbb{E}_{\tau\sim\pi_\theta}[\sum_t\gamma^t r_t]$, with no gradients or structural knowledge of $J$; one evaluation is a full rollout. Classical optimizers carry explicit state (a distribution, covariance, population); JEPA-PS is instead a stateless \emph{in-context} function reading history $\mathcal{H}_t=\{(\theta_i,r_i)\}_{i\le t}$ from a sliding buffer of capacity $L$ and emitting
\begin{equation}
\hat\theta_{t+1}=f_\phi(\mathcal{H}_t,r^\ast),\qquad r^\ast=\mathrm{pct}_{80}(\{r_i\}_{i\le t}),
\label{eq:fpol}
\end{equation}
where $r^\ast$ is a target-return conditioning signal. This isolates the two inputs we probe: the history, a set we can subset, reorder, and permute, and the reward encoding, which we can rank, rescale, permute, or corrupt. Figure~\ref{fig:arch} overviews the resulting architecture.

\paragraph{Pretraining.} We collect $(\theta,J(\theta))$ pairs from random $\theta\sim\mathcal{N}(0,I)$ on source environments spanning balance, locomotion, and control; the corpus holds only scalar returns and raw parameter vectors (no task descriptions, observations, or reward shaping), returns percentile-normalized. Windows of length $L_{\mathrm{ctx}}\in[4,20]$ slide over the stored pairs. A Transformer context encoder $g_\phi$ \citep{vaswani2017attention} maps a window of mean-normalized $(\bar\theta_i,r_i)$ tokens to a latent $s_x\in\mathbb{R}^{d_m}$; a predictor $p_\psi$ maps $(s_x,r^\ast)$ to $\hat s_y$; a decoder $h_\xi$ maps latents to parameters, with Gaussian noise $\epsilon\sim\mathcal{N}(0,\sigma_n^2I)$ added at inference for exploration. Encoder and predictor train against a stop-gradient target from an EMA copy $g_{\bar\phi}$ applied to a window including the next pair:
\begin{equation}
\mathcal{L}_{\mathrm{JEPA}}=\big\|\hat s_y-\mathrm{sg}(s_y)\big\|_2^2,\qquad s_y=g_{\bar\phi}\big(\{(\bar\theta_i^+,r_i^+)\}\big).
\label{eq:loss}
\end{equation}
In plain terms, the model predicts what the next latent summary will look like, not the reward itself. Two properties matter for what follows. \emph{(i) The loss never predicts reward:} shuffling the $\theta$--$r$ correspondence during pretraining leaves downstream performance unchanged (Appendix~\ref{app:shuffle}), yet reward information is used at test time (Section~\ref{sec:analysis}), by a mechanism the objective never supervised. \emph{(ii) The decoder receives no gradient:} $h_\xi$ does not appear in Eq.~\ref{eq:loss} and is a fixed random projection in the default configuration (measured gradient norms: encoder 0.248, predictor 0.453, decoder 0.000; Appendix~\ref{app:decoder}), so the default system is \emph{latent-guided}.

\paragraph{Inference and population wrapper.} Algorithm~\ref{alg:jepa} is the single-agent loop. Since history can collapse onto one region on harder tasks, we optionally wrap $K$ agents with independent warm-starts, migration of the global best every 30 iterations, and isolation phases suspending migration for 6 steps every 60 iterations \citep{whitley1999island,kennedy1995particle} -- Swarm-JEPA. Section~\ref{sec:transfer} separately introduces a restart-on-stagnation wrapper, reinitializing one agent's history rather than running several in parallel. Every method here, learned or classical, is charged the same \emph{total} rollouts (fixed-total accounting; Appendix~\ref{app:budgetfair}).

\begin{algorithm}[t]
\caption{JEPA in-context policy optimization (single agent)}
\label{alg:jepa}
\begin{algorithmic}[1]
\Require pretrained $g_\phi,p_\psi,h_\xi$; environment $E$; budget $T$; capacity $L$; warm-start length $L_{\mathrm{warm}}$
\State $\mathcal{H}\gets\emptyset$; warm-start with $L_{\mathrm{warm}}$ random rollouts
\For{$t=L_{\mathrm{warm}}{+}1,\ldots,T$}
  \State $\mu,\sigma\gets\mathrm{mean},\mathrm{std}$ of $\{\theta_i\}$ over $\mathcal{H}$;\quad $r^\ast\gets\mathrm{pct}_{80}(\{r_i\})$
  \State $\hat s_y\gets p_\psi\big(g_\phi(\{((\theta_i{-}\mu)/\sigma,\ r_i)\}),\ r^\ast\big)$
  \State $\hat\theta_t\gets h_\xi(\hat s_y)\cdot\sigma+\mu+\epsilon,\qquad \epsilon\sim\mathcal{N}(0,\sigma_n^2 I)$
  \State $r_t\gets J(\hat\theta_t)$;\quad $\mathcal{H}\gets\mathcal{H}\cup\{(\hat\theta_t,r_t)\}$
\EndFor
\State \Return $\arg\max_{(\theta,r)\in\mathcal{H}} r$
\end{algorithmic}
\end{algorithm}

% ═══════════════════════════════════════════════
\section{Experimental Setup}
\label{sec:setup}
% ═══════════════════════════════════════════════
\paragraph{Environments and baselines.} We use 19 control environments \citep{brockman2016openai,todorov2012mujoco}: balance (CartPole, InvertedPendulum, InvertedDoublePendulum), locomotion (Swimmer, Hopper, Walker2d, Reacher), continuous/discrete control (MountainCarContinuous, MountainCar), discrete navigation (FrozenLake, CliffWalking), and six custom/externally-sourced tasks (Maze, Nim, Navigation, Pong, BallLanding), linear policies $\pi_\theta(s)=Ws+b$, $d\in[3,108]$ \citep{mania2018simple}. Eight were added to broaden coverage beyond continuous control; Appendix~\ref{app:extended} gives provenance, confirms dimensionality against source. We compare against 21 optimizers: classical (CEM, CMA-ES, PSO, BO, OpenAI-ES, $(\mu{+}\lambda)$-ES, Tabu, random), strong modern baselines (TuRBO, SMAC, TPE, IPOP-/BIPOP-CMA-ES, L-BFGS-B), policy-search/amortized methods (ARS, CEM-RL, CERL, PES, PABBO), an OPRO-style LLM optimizer (Qwen2.5-7B, \citep{yang2024large}), and an LLAMBO adaptation \citep{liu2024llambo} (Appendix~\ref{app:llambo}). Full grids: Appendix~\ref{app:baselines},~\ref{app:strong}.

\paragraph{Protocol.} Every method receives $T{=}400$ total evaluations, $n_{\mathrm{ep}}{=}10$ rollouts each, 3--5 seeds, on the 11 original environments; we report mean best reward $\pm$ std. The eight extended environments (Appendix~\ref{app:extended}) use a timestep-matched budget ($T{=}2$,000--20,000, $n_{\mathrm{ep}}{=}5$, $n{=}10$ seeds) with classical/policy-search baselines only. Swarm-JEPA uses $K{=}3$, $\sigma_n{=}0.3$. Section~\ref{sec:analysis}'s ablations use the reduced-budget protocol of Appendix~\ref{app:forget}; Section~\ref{sec:transfer} uses its own pre-registered split. A standardized 10-seed comparison is in Appendix~\ref{app:standardized}.

\paragraph{Quantifying memory and landscape.} We summarize each environment's trajectory use with two scalars derived from inference-time ablations: \emph{history dependence}, the fractional gain from using the full history rather than the best-seen point, and \emph{ordering sensitivity}, the fractional loss when history rewards are randomly permuted:
\begin{equation}
\HD=\frac{R_{\mathrm{full}}-R_{\mathrm{best\text{-}only}}}{|R_{\mathrm{full}}|},\qquad
\OS=\frac{R_{\mathrm{orig}}-R_{\mathrm{perm}}}{|R_{\mathrm{orig}}|}.
\label{eq:hd}
\end{equation}
Both are zero when the optimizer ignores the ablated information. The landscape descriptor is the fitness--distance correlation, $\FDC=\mathrm{corr}\big(J(\theta),-\|\theta-\theta^\ast\|\big)$ \citep{jones1995fitness}; since $\theta^\ast$ is unknown we use the best probe sample as proxy, computed from 150--300 random evaluations, and treat $\FDC$ as an ordinal smoothness descriptor (sample-size sensitivity in Appendix~\ref{app:fdcsens}). Five further descriptors (ACL, basin size, smoothness, neutrality, information content, dispersion) are defined in Appendix~\ref{app:fdcsens}.

% ═══════════════════════════════════════════════
\section{JEPA-PS as an Optimizer}
\label{sec:results}
% ═══════════════════════════════════════════════
Section~\ref{sec:analysis} examines what information JEPA-PS uses, and Section~\ref{sec:transfer} tests what its pretraining transfers.

\paragraph{Competitive with classical search, but not consistently better.}
\emph{Swarm-JEPA is competitive with tuned classical search on several tasks, but it does not consistently outperform it.}
Swarm-JEPA ($K{=}8$) reaches 416.1 on CartPole, 293.1 on Swimmer, and 1039.7 on Hopper. These scores exceed CEM's respective scores of 360.9, 250.8, and 620.2 (Appendix~\ref{app:baselines}, Table~\ref{tab:main}). IPOP-CMA-ES and ARS solve CartPole and reach approximately 352 on Swimmer, while TuRBO reaches 358.0 on Swimmer. On Hopper, TuRBO, BIPOP-CMA-ES, and TPE finish within approximately 40 points of Swarm-JEPA (Table~\ref{tab:strong_grid}). On InvertedPendulum, JEPA-PS reaches only 27.9, compared with IPOP-CMA-ES's 890.0. Diagnostics in Appendix~\ref{app:invpend_diag} do not indicate an encoder or decoder failure. A 300-sample probe finds a best reward of only approximately 58, about 15 times below IPOP-CMA-ES's score. These results are consistent with difficulty finding the high-reward basin, rather than simply failing to use a trajectory once that basin is found. The evaluated JEPA-PS configuration has no restart mechanism; Section~\ref{sec:transfer} tests whether adding one helps. LLM comparisons appear in Section~\ref{sec:llm}, and the dedicated CMA-ES battery appears in Appendix~\ref{app:cmaes_battery}.

\paragraph{Eight additional environments.}
Table~\ref{tab:extended_baselines} in Appendix~\ref{app:extended} reports ARS, CEM, CMA-ES, and PES results across eight additional domains. Pong, InvDblPendulum, MountainCar(D), and Navigation show variation across methods, while FrozenLake and Nim yield weak performance.

\emph{On CliffWalking, all four methods converge to the same cliff-fall outcome.}
The result has zero variance across ten seeds. The diagnostics in Appendix~\ref{app:extended} attribute this behavior to the landscape under linear policies rather than an implementation error. Random policies concentrate in two outcome classes: an immediate cliff fall or full-length wandering. Almost no intermediate outcomes occur, and every evaluated method converges to the cliff-fall outcome.

\paragraph{Reward-relevant structure in the latent space.}
\emph{The latent space contains reward-relevant information, but its advantage over a sampling control depends on the protocol.}
In the ten-seed comparison in Table~\ref{tab:null_model}, a null-proposer sampling method reaches 302.1 on CartPole and 151.9 on Swimmer. JEPA-PS reaches 410.9 and 293.1, respectively. However, a more closely matched Swimmer comparison, using shared warm starts and noise seeding, finds that the null model matches or exceeds JEPA-PS (Appendix~\ref{app:decoder}). Separately, latent norm correlates with reward at 0.507, compared with 0.015 for a random projection. This correlation does not, by itself, show that the learned proposer uses that structure better than a matched sampling control.

% ═══════════════════════════════════════════════
\section{JEPA-PS versus LLM-Based Optimizers}
\label{sec:llm}
% ═══════════════════════════════════════════════
We compare JEPA-PS with several LLM-based optimization systems: the OPRO-style and LLAMBO baselines in Table~\ref{tab:main}, an independent five-seed comparison with two LLM agents, and restart-wrapped LLM evaluations on nominal and perturbed CartPole. These experiments use different implementations and search configurations. We interpret their scores within their respective protocols.

\paragraph{Single-call LLM optimization and LLAMBO.}
\emph{Swarm-JEPA performs better than the single-call LLM optimizer on CartPole and Swimmer, but not on every task.}
Swarm-JEPA reaches 416.1 on CartPole and 293.1 on Swimmer, compared with 186.0 and 140.3 for the LLM optimizer. The methods are comparable on Hopper (1039.7 versus 1026.7), while the LLM performs better on InvertedPendulum (356.0 versus 27.9). JEPA-PS makes no language-model call at search time and requires no text parsing. Its measured proposal latency is 2.61\,ms, compared with approximately 200\,ms for the 7B LLM, a roughly 77-fold difference (Table~\ref{tab:costfull}). Classical updates are cheaper still: CEM takes 0.024\,ms and CMA-ES takes 0.053\,ms in the measured configurations. JEPA-PS uses a fixed-size latent representation, and its pretraining takes a few GPU-minutes (Appendix~\ref{app:pretraining_cost}). LLAMBO \citep{liu2024llambo} adds an LLM-generated candidate pool and a surrogate-ranking step on the same backend (Appendix~\ref{app:llambo}). Many ranking calls fall back to unranked defaults after parsing failures. This experiment therefore measures LLAMBO with an unreliable ranking stage, not its performance under consistently successful ranking.

\paragraph{An independent comparison with LLM agents.}
\emph{Both JEPA variants have higher mean scores than the two LLM agents, but plain CEM performs best.}
A separate five-seed CartPole experiment compares JEPA-PS and Swarm-JEPA with two Qwen2.5:7B agents, ProPS and the retrieval-augmented ProPS-Redundant, alongside CEM and random search at a matched budget (Table~\ref{tab:compare_methods}; per-seed and per-task breakdowns, including the MountainCar robustness sweep, are in Appendix~\ref{app:props}).

\begin{table}[h]
\centering
\caption{Independent five-seed CartPole comparison at a matched budget. Scores report mean best reward $\pm$ standard deviation.}
\label{tab:compare_methods}
\small
\begin{tabular}{lr}
\toprule
Method & Mean $\pm$ std\\
\midrule
CEM Baseline & $500.00{\pm}0.00$\\
JEPA (ours) & $365.79{\pm}182.71$\\
Swarm-JEPA (ours) & $223.73{\pm}114.07$\\
ProPS-Redundant (Qwen2.5:7B) & $170.53{\pm}74.25$\\
ProPS (Qwen2.5:7B) & $103.01{\pm}43.47$\\
Random Search & $60.52{\pm}14.49$\\
\bottomrule
\end{tabular}
\end{table}

\paragraph{Restart-wrapped LLM optimization on nominal CartPole.}
\emph{The restart-wrapped LLM nearly solves nominal CartPole, but this score includes random fallback after parsing failures.}
Restarting after 40 consecutive non-improving evaluations raises the score to $499.44{\pm}1.68$ over ten seeds. The parse-failure rate is 0.387, or 38.7\% (Table~\ref{tab:llm_heldout}). The score therefore measures the complete system: valid LLM proposals, random-noise fallback, and the restart wrapper. The per-evaluation trace contains long stretches at reward 500 interrupted by drops to approximately 9, recurring on roughly the restart timescale.

\paragraph{What the comparisons explain.}
The weaker LLM-agent scores cannot be attributed solely to complete generation failures. ProPS and ProPS-Redundant report a $0.0\%$ complete-failure rate across their five-seed runs. However, complete failure and per-call parsing failure are different measures. Across the evaluated OPRO-style, LLAMBO, and ProPS configurations, LLM-based search does not consistently match JEPA-PS on CartPole or Swimmer. Restarting substantially improves the complete LLM system on CartPole. The present experiments do not isolate how much of this gain comes from valid LLM proposals, fallback sampling, or restarting. JEPA-PS is trained on continuous $(\theta,r)$ trajectory structure, whereas the general LLM must infer a search direction from serialized examples without the same task-specific objective. This account is compatible with \citep{ferreira2026centaur} and the modest LLAMBO gains observed here, but it remains an interpretation rather than an experimentally established mechanism.

\paragraph{Held-out CartPole physics variants.}
\emph{The restart-wrapped LLM maintains high scores on the held-out physics variants in this experiment.}
We evaluate ten seeds per variant with a budget of 400 evaluations and a restart after 40 non-improving evaluations. Table~\ref{tab:llm_heldout} includes the three training variants and four held-out variants from Appendix~\ref{app:cmaes_battery}. These held-outs include both single-factor and combined perturbations. They are distinct from the four transfer-test variants in Section~\ref{sec:transfer}. Mean best reward ranges from 480.92 to 500.00 on the training variants and from 474.06 to 500.00 on the held-out variants. All four held-out variants have zero parsing failures, while \texttt{force\_x0.5} has a failure rate of 0.045. Nominal CartPole and \texttt{gravity\_x2} have higher failure rates of 0.387 and 0.569, respectively. These results show no obvious held-out performance deficit for the complete restart-wrapped system on this perturbation family. They do not separate the contribution of the LLM's proposal rule from that of the wrapper.

\begin{table}[h]
\centering
\caption{Restart-wrapped LLM optimization on training and held-out CartPole physics variants: ten seeds, budget 400, and restart after 40 non-improving evaluations. Scores report mean best reward $\pm$ standard deviation. Parse-failure rates are fractions of calls; failed calls use random-noise fallback.}
\label{tab:llm_heldout}
\small
\begin{tabular}{llrr}
\toprule
Variant & Split & Mean $\pm$ std & Parse-failure rate\\
\midrule
nominal & TRAIN & $499.44{\pm}1.68$ & $0.387$\\
gravity\_x2 & TRAIN & $500.00{\pm}0.00$ & $0.569$\\
force\_x0.5 & TRAIN & $480.92{\pm}51.04$ & $0.045$\\
gravity\_x1.5+length\_x1.3 & HELD-OUT & $474.06{\pm}57.59$ & $0.000$\\
masspole\_x2 & HELD-OUT & $487.12{\pm}38.24$ & $0.000$\\
masscart\_x0.5+force\_x1.5 & HELD-OUT & $500.00{\pm}0.00$ & $0.000$\\
length\_x0.5+force\_x2 & HELD-OUT & $500.00{\pm}0.00$ & $0.000$\\
\bottomrule
\end{tabular}
\end{table}

% ═══════════════════════════════════════════════
\section{What Does the Optimizer Use?}
\label{sec:analysis}
% ═══════════════════════════════════════════════
We now examine JEPA-PS through controlled interventions. These experiments change the history presented to the model while leaving the true environment evaluations unchanged. The protocols use different budgets and seed configurations, so comparisons are made within each experiment.

\begin{table}[t]
\caption{Inference-time history interventions. Panel A uses CartPole with five seeds and 200 iterations; $\dagger$ denotes three seeds. Panel B uses the reduced-budget protocol in Appendix~\ref{app:forget}. The panels use different protocols, so their reward-permutation effects are not directly interchangeable.}
\label{tab:ablation_main}
\small
\begin{minipage}[t]{0.5\textwidth}
\centering
A\\[2pt]
\begin{tabular}{lrr}
\toprule
Intervention & Mean best & $\Delta$\\
\midrule
Control (full history) & $410.9$ & ---\\
Best-only (single point) & $319.2$ & $-22.3\%$\\
Remove best-seen point$^\dagger$ & $285.1$ & $-30.6\%$\\
Permute parameter dims. & $329.7$ & $-19.8\%$\\
Shuffle reward values & $341.1$ & $-17.0\%$\\
Reverse temporal order & $354.7$ & $-13.6\%$\\
Shuffle temporal order$^\dagger$ & $410.9$ & $+0.0\%$\\
\bottomrule
\end{tabular}
\end{minipage}\hfill
\begin{minipage}[t]{0.48\textwidth}
\centering
B\\[2pt]
\setlength{\tabcolsep}{4pt}
\begin{tabular}{lrrr}
\toprule
Rewards & CartPole & Swimmer & MtnCar(C)\\
\midrule
Original & $337.9$ & $263.4$ & $-28.41$\\
Ranked & $338.3$ & $262.9$ & $-28.39$\\
Monotone (log) & $337.7$ & $263.1$ & $-28.42$\\
Permuted & $212.8$ & $245.7$ & $-28.40$\\
\midrule
$\OS$ & $0.370$ & $0.068$ & $-0.000$\\
\bottomrule
\end{tabular}
\end{minipage}
\end{table}

\paragraph{History content matters more than buffer capacity in the tested setting.}
\emph{Changing what the model remembers matters more than increasing the buffer size in the low-budget experiment.}
Sweeping buffer capacity $L$ from 4 to 400 leaves best reward unchanged under the fixed low-budget protocol of Table~\ref{tab:history_ablations}. This result remains to be tested at longer budgets, such as $T{=}400$. In the history-content comparison reported in Table~\ref{tab:history_ablations}, full history reaches 410.9, best-only reaches 325.5, and recent-8 reaches 333.0. CEM reaches 360.9 in the baseline comparison, although that comparison changes the search rule as well as its memory.

\paragraph{Several parts of the history affect performance.}
\emph{JEPA-PS uses more than the best point in its history.}
In Table~\ref{tab:ablation_main}A, performance falls when the history is reduced to its best point ($-22.3\%$), parameter dimensions are permuted ($-19.8\%$), reward values are shuffled ($-17.0\%$), or temporal order is reversed ($-13.6\%$). These results show sensitivity to trajectory content, coordinate identity, parameter--reward associations, and some changes in ordering. Randomly shuffling temporal order has no measured cost in the three-seed experiment, whereas reversing it does. This difference does not uniquely identify a temporal-direction mechanism. The pretraining-time shuffle in Appendix~\ref{app:shuffle} is a separate intervention and also has no measured performance effect.

\paragraph{Reward ordering matters more than the tested reward scales.}
\emph{The tested changes in reward scale have little effect, but shuffling rewards can hurt performance.}
Original rewards, rank-transformed rewards, and a monotone logarithmic transformation produce nearly identical scores in Table~\ref{tab:ablation_main}B. This is consistent with reliance on relative reward ordering, as in rank-based archives \citep{alansari2026sparrow}. It does not establish invariance to every monotone transformation. Permuting rewards reduces performance by approximately 37\% on CartPole and 6.8\% on Swimmer, while producing essentially no change on MountainCar. Section~\ref{sec:landscape} examines this variation.

\paragraph{History rewards matter more than the target-return input.}
\emph{Removing rewards from the history hurts performance; changing the target-return input does not.}
In Eq.~\ref{eq:fpol}, the predictor receives
$r^\ast=\mathrm{pct}_{80}(\{r_i\}_{i\le t})$
as a target-return input. This value is separate from the per-step rewards stored in the history. Setting $r^\ast$ to zero, a random value, or the worst observed reward produces no measured change in final CartPole performance. Percentiles from 50 to 99 also produce identical results (Appendix~\ref{app:rstar}). In contrast, zeroing the history rewards reduces performance by 34.8\%. Changing $r^\ast$ does alter the proposal distribution. On Swimmer, the reported shifts are 5--9 standard deviations, with MMD values of 1.6--1.8. The model therefore responds to this input, but the response provides no measured final-reward benefit in these experiments.

\paragraph{Pretraining provides a useful latent geometry.}
\emph{Classical search can use the pretrained latent space without relying on the learned proposal rule.}
Latent-CEM matches Swarm-JEPA on CartPole and Swimmer in the reported comparison (Table~\ref{tab:main}). It also improves over raw-space CEM by 41.6\% in the three-seed experiment of Appendix~\ref{app:latentcem} and by 21\% in a separate five-seed experiment. Reconstruction fine-tuning of $h_\xi$ reduces CartPole reward from 410.9 to 293.7 ($-28.5\%$), and end-to-end retraining reduces it to 269.0 ($-34.5\%$; Appendix~\ref{app:decoder}). Across ten random linear decoders applied to the same pretrained latents, the between-decoder standard deviation is 26.7, compared with within-decoder seed variation of 178.2. Performance is relatively insensitive to the particular fixed linear readout in this experiment. These findings favor a representational contribution from pretraining over a uniquely effective decoder-specific proposal mechanism. They do not imply that all decoder changes are harmless: the fine-tuning interventions reduce performance substantially. The learned proposal direction is also nearly orthogonal to six classical update directions, with cosine similarities of 0.003--0.015 (Appendix~\ref{app:mech}). This directional difference does not, by itself, establish a performance advantage.

% ═══════════════════════════════════════════════
\section{Does Landscape Structure Predict the Memory Requirement?}
\label{sec:landscape}
% ═══════════════════════════════════════════════
Section~\ref{sec:analysis} shows that the benefit of history varies across environments. We now ask whether that benefit is associated with an independently measured property of the search landscape, and how much evidence supports the association.

\paragraph{History helps by different amounts across environments.}
\emph{History helps most on CartPole and has little measured benefit on several other tasks.}
Table~\ref{tab:hdmain} reports history dependence $\HD$ and ordering sensitivity $\OS$ from Eq.~\ref{eq:hd}, together with fitness--distance correlation $\FDC$, for seven environments with JEPA checkpoints. History dependence ranges from 0.218 on CartPole to approximately zero on MountainCar, Walker2d, and Hopper. In the cross-environment history-dependence protocol of Table~\ref{tab:hd}, CartPole reaches 410.9 with full history and 321.4 with best-only history. This best-only score is separate from the score of 319.2 in the five-seed, 200-iteration intervention experiment of Table~\ref{tab:ablation_main}A and the score of 325.5 in the history-content comparison of Table~\ref{tab:history_ablations}. On MountainCar, the full-history and best-only scores in the $\HD$ evaluation are nearly identical: $-28.41$ and $-28.51$.

\begin{table}[t]
\caption{Fitness--distance correlation ($\FDC$), history dependence ($\HD$), and reward-ordering sensitivity ($\OS$). The reported Pearson correlation between $\HD$ and $\FDC$ is 0.738 across seven environments, with a bootstrap 95\% confidence interval excluding zero. The original $\OS$ correlation uses five environments, excluding InvertedPendulum and Hopper. The all-seven $\OS$ calculation includes their displayed zero entries. The all-seven $\OS$ and Spearman correlations are computed from the rounded entries shown.}
\label{tab:hdmain}
\small
\begin{center}
\begin{tabular}{lrrr}
\toprule
Environment & $\FDC$ & $\HD$ & $\OS$\\
\midrule
CartPole & $0.661$ & $0.218$ & $0.370$\\
InvPendulum & $0.548$ & $0.050$ & $0.000$\\
Swimmer & $0.482$ & $0.044$ & $0.068$\\
Hopper & $0.439$ & $-0.006$ & $0.000$\\
Reacher & $0.401$ & $0.074$ & $0.003$\\
Walker2d & $0.326$ & $0.000$ & $0.000$\\
MountainCar(C) & $0.153$ & $0.004$ & $-0.000$\\
\midrule
Pearson $r$ with $\FDC$, all 7 & & $0.738$ & $0.66$\\
Pearson $r$ with $\FDC$, 5-env.\ subset & & --- & $0.832$\\
Spearman $\rho$ with $\FDC$, all 7 & & $0.571$ & $0.571$\\
\bottomrule
\end{tabular}
\end{center}
\end{table}

\begin{figure}[t]
\centering
\includegraphics[width=0.92\linewidth,height=5cm]{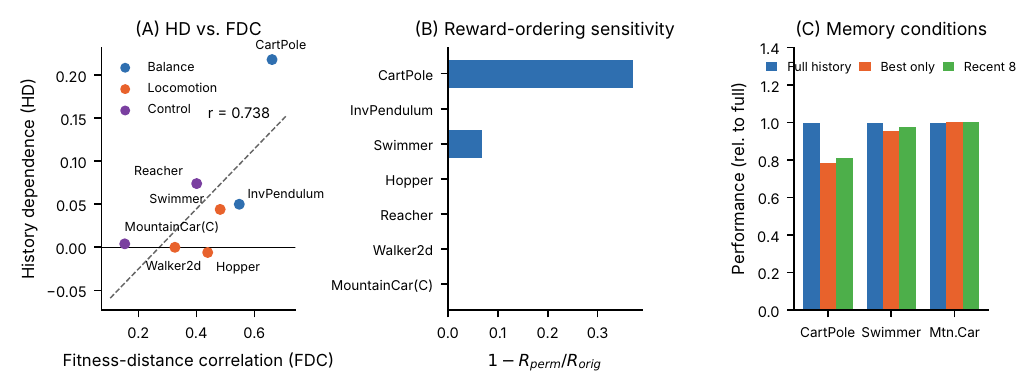}
\caption{Landscape structure and measured history benefits. (A) History dependence versus fitness--distance correlation across seven environments. (B) Reward-ordering sensitivity versus FDC. The tested monotone reward transformations have little effect, while reward permutation has an environment-dependent effect. (C) Memory-condition performance normalized to the full-history baseline.}
\label{fig:landscape}
\end{figure}

\paragraph{History dependence is associated with FDC.}
\emph{Higher FDC is associated with a larger history benefit in this small sample.}
Across seven environments, the Pearson correlation between $\HD$ and $\FDC$ is 0.738 (Table~\ref{tab:hdmain}, Figure~\ref{fig:landscape}). The correlation between $\OS$ and $\FDC$ is 0.832 in the five-environment analysis. These probes change different aspects of the history, but they are not independent replications. They use overlapping environments and checkpoints, and both associations are strongly influenced by CartPole. On synthetic NK landscapes, FDC decreases monotonically from 0.389 to 0.028 as epistasis increases (Appendix~\ref{app:fdcsens}). This supports its use as a landscape descriptor in that controlled setting. It does not independently establish a link between FDC and memory.

\paragraph{Limits of the evidence.}
The Spearman correlation between $\HD$ and $\FDC$ is 0.571, below the two-tailed significance threshold at $n{=}7$. Removing CartPole reduces Pearson correlation to approximately 0.50 and Spearman correlation to approximately 0.31. The $\OS$ correlation of 0.832 applies only to the five-environment subset; the all-seven calculation using the displayed entries gives approximately 0.66. Conditioning on ACL changes the partial correlation to $-0.487$. With only seven environments and substantial FDC--ACL collinearity, this sign change does not support a stable conditional interpretation (Appendix~\ref{app:fdcsens}). History dependence is computed from best-over-history reward. A near-zero value may mean that extra history is unnecessary, but it may also mean that neither condition improves beyond initialization. This distinction matters on floor-limited tasks such as MountainCar and in basin-discovery failures such as InvertedPendulum. The record-statistics analysis in Appendix~\ref{sec:robustness} finds more warm-up-determined best scores than its null model predicts, with $p<10^{-4}$ in each environment tested. Initialization dominance is therefore a substantive confound for best-over-history metrics. Section~\ref{sec:transfer} also measures post-warm-up gain to address it.

\paragraph{Interpretation.}
\emph{The results suggest a link between landscape structure and history use, but they do not establish a general rule.}
CartPole has both the highest FDC and the largest measured history benefit. Lower-FDC environments show less benefit. However, the experiments do not fully separate a sufficient best point from a failure to make progress. The checkpoints are also environment-specific, although they share an architecture and training recipe (Appendix~\ref{app:impl}). The comparison therefore concerns that shared recipe, not a single weight vector used across tasks.

% ═══════════════════════════════════════════════
\section{Does Pretraining Transfer, or Memorize?}
\label{sec:transfer}
% ═══════════════════════════════════════════════
Every checkpoint in Sections~\ref{sec:results}--\ref{sec:landscape} is pretrained and evaluated within a single environment. These results alone do not establish transfer beyond the pretraining dynamics. The tests separate three questions: whether pretrained weights help on unseen dynamics, whether that help depends on the supplied context, and whether the resulting optimizer competes with classical search.

\paragraph{Part 1: transfer to held-out physics variants.}
\emph{Pretraining helps on CartPole physics variants that were not used in training.}
We pretrain an encoder and predictor on nominal CartPole, gravity$\times$2, and force$\times$0.5. The control uses the same decoder but a reinitialized encoder and predictor. We evaluate both models on these training variants and on two held-outs: gravity$\times$0.5 and length$\times$2. Across 100 paired seeds, the pretrained model exceeds the control on every training variant ($p<10^{-5}$) and both held-outs ($p<10^{-8}$). The held-out differences are 104--110 reward points, compared with a mean training difference of approximately 83 points. This is within-environment physics transfer, similar to domain randomization \citep{tobin2017domain,peng2018sim}, not cross-task generalization. Parts~4 and~6 compare optimization performance with classical search. Appendix~\ref{app:cmaes_battery} provides the separate CMA-ES battery.

\paragraph{Part 2: larger, joint physics perturbations.}
\emph{The pretraining advantage persists when two physical constants change together.}
Training changes one constant at a time, using multipliers of 2.0 or 0.5. We test two held-out variants with joint changes and unseen multipliers: gravity$\times$3.0 plus length$\times$1.5 (EXTREME-A), and gravity$\times$0.33 plus force$\times$2.0 (EXTREME-B). Without retraining, the pretrained model exceeds the untrained control on both variants over 100 seeds. The difference is $+50.6$ on EXTREME-A ($p{=}2.0{\times}10^{-5}$), approximately 61\% of Part~1's mean training effect. It is $+90.2$ on EXTREME-B ($p{=}2.9{\times}10^{-8}$), approximately 108\%.

\paragraph{Part 3: matched versus mismatched context.}
\emph{The context-swap test does not detect a reliable benefit from matched warm-up rewards.}
Parts~1 and~2 show that pretrained weights help. They do not show whether that help comes from adapting to the current variant through the history. To test this, we keep the warm-up parameter draws fixed but replace their rewards with rewards from a different physics variant. If variant-specific warm-up information helps, matched histories should perform better than mismatched histories. A run's best reward can be determined entirely during warm-up, which can hide differences in later search. We therefore compare post-warm-up gain. Across 100 seeds, the matched-minus-mismatched difference is $+3.4$ on gravity$\times$0.5 ($p{=}0.85$; 72 informative seeds) and $+24.0$ on length$\times$2 ($p{=}0.37$; 87 informative seeds).

\paragraph{Part 4: classical search on the two Part~1 held-outs.}
\emph{On both Part~1 held-outs, ARS and CEM outperform unwrapped JEPA-PS.}
We run ARS and CEM from scratch at $T{=}400$ on gravity$\times$0.5 and length$\times$2. Both reach $500.0{\pm}0.0$ across 20 seeds on each variant. Unwrapped JEPA-PS reaches $437.5{\pm}139.2$ and $319.9{\pm}156.4$, respectively. This comparison shows that transfer relative to an untrained control does not provide an advantage over classical search here. Part~6 reports the broader four-variant comparison, adding EXTREME-A and EXTREME-B. Appendix~\ref{app:cmaes_battery} compares against CMA-ES.

\paragraph{Part 5: selecting an inference-time intervention.}
\emph{Inference-time interventions can greatly improve success, but that gain alone does not show a benefit from pretraining.}
We evaluate 17 interventions on three validation variants that are separate from the test set: force$\times$1.5, gravity$\times$1.5, and length$\times$0.6. The interventions include longer warm-ups, noise schedules, restart-on-stagnation, multi-start search, selection variants, and combinations. Several configurations raise pooled success, defined as reward $\geq490$, from 0.38 to as high as 0.99. A wrapper could also improve an untrained network, so we run the corresponding untrained controls. We use the pretrained-versus-untrained comparison when selecting the final configuration. Restarting after 40 non-improving evaluations is carried forward to Part~6. Full per-configuration numbers and the pre-registered attribution rule are in Appendix~\ref{app:wrapper}.

\paragraph{Part 6: final evaluation on four held-out variants.}
\emph{Restarting helps both models, and the pretrained model remains ahead of the untrained control.}
The final evaluation uses 100 seeds on four held-out variants: gravity$\times$0.5, length$\times$2, EXTREME-A, and EXTREME-B. Restart-after-40 raises pooled pretrained success from 0.47 to 0.84. It raises pooled untrained-control success from 0.14 to 0.66. Thus, the wrapper contributes substantially, while pretrained weights retain an advantage under the same wrapper.

\emph{Wrapped JEPA-PS matches the classical success rate on two variants, but not on all four.}
Its success rate is 1.00 on gravity$\times$0.5 and EXTREME-B, 0.81 on length$\times$2, and 0.56 on EXTREME-A. The four-variant ARS/CEM comparison gives a success rate of 1.00 for both classical methods on every variant. The restart intervention narrows the overall gap but does not close it.

\paragraph{Part 7: broader training curricula.}
\emph{Neither tested curriculum expansion produces a reliable improvement.}
First, we broaden pretraining by sampling gravity, force, and length log-uniformly over $[0.7,1.5]$, excluding test multipliers. The broader curriculum is directionally ahead on all three Part~5 validation variants. However, the pooled success-rate gain of $+0.06$ falls below the 0.10 criterion, and no individual variant shows a significant improvement. Second, we pretrain on a mixture of CartPole and synthetic NK landscapes with $K{=}0,2,4,8$. Zero-shot evaluation on Rastrigin gives $-49.5{\pm}8.1$ over 20 seeds. This does not show a detectable improvement over the narrow-curriculum or BBOB-only results, both approximately $-53.0$ (Appendix~\ref{app:bbob}). All remain well below CEM's $-16.2$.

\paragraph{Summary of the transfer investigation.}
\emph{Pretraining transfers across CartPole dynamics, but it does not establish a practical advantage over classical search.}
The pretrained model exceeds its untrained control on unseen physics variants, including large joint perturbations. However, the limited context-swap test does not identify variant-specific warm-up conditioning as the cause. The best tested intervention also does not consistently match ARS or CEM, and the curriculum expansions provide no reliable additional gain. Appendix~\ref{app:cmaes_battery} extends the investigation with a direct six-part comparison against CMA-ES.

% ═══════════════════════════════════════════════
\section{Discussion}
\label{sec:discussion}
% ═══════════════════════════════════════════════
\paragraph{Memory benefits depend on the problem.}
\emph{The same architecture and training recipe can show very different benefits from history across tasks.}
History interventions have a large effect on CartPole and little measured effect on several other environments. This suggests that useful memory depends on the search problem, not only on the optimizer's design. The association with FDC offers one possible explanation. However, the small sample, CartPole's influence, and warm-up-dominated runs limit that conclusion. Little measured history benefit on a difficult task does not prove that the best point is sufficient. It may instead mean that neither condition makes progress. The results therefore motivate landscape-aware studies of memory, rather than a single prescription for how much history every optimizer should retain.

\paragraph{What pretraining provides.}
\emph{The strongest evidence is for a useful representation, not a uniquely effective learned proposal rule.}
Latent-CEM uses the pretrained geometry without relying on the learned decoder and matches Swarm-JEPA in the reported comparisons. The decoder experiments support the same interpretation. The physics-transfer results further show that the benefit of pretrained weights is not confined to the exact training dynamics. The practical and mechanistic limits remain clear. The context-swap probe does not establish how variant-specific history supports transfer. Restarting improves both pretrained and untrained systems, and JEPA-PS still does not consistently match ARS or CEM across the transfer test set. The direct CMA-ES battery also finds a representational signal without a performance advantage at the evaluated budgets (Appendix~\ref{app:cmaes_battery}).

% \paragraph{Limitations.}The link between landscape and memory is based on seven environments, with CartPole contributing much of the leverage. The results are sensitive to subset choice, while warm-up dominance and floor effects confound the hardest tasks (Section~\ref{sec:landscape}). The transfer investigation (Section~\ref{sec:transfer}) is confined to CartPole physics variants; we have not repeated it on a harder or structurally different environment, the most important follow-up given the mixed results already obtained. We use linear policies and environment-specific checkpoints; the eight extended environments (Appendix~\ref{app:extended}) have no JEPA-PS checkpoint, and Maze has no complete baseline. Our LLAMBO comparison (Appendix~\ref{app:llambo}) is a single-environment, $n{=}3$ pilot with an imperfect parser, and the multi-variant LLM generalization test (Section~\ref{sec:llm}) is unresolved pending an infrastructure fix. The optimizer does not adapt to mid-run objective changes (Appendix~\ref{sec:robustness}) and fails on InvertedPendulum. Classical updates, including CMA-ES across the full battery of Appendix~\ref{app:cmaes_battery}, are cheaper per step and, on current evidence, never worse in final performance.

% ═══════════════════════════════════════════════
\section{Conclusion}
% ═══════════════════════════════════════════════
We asked what a learned optimizer actually uses of its history, and built a small, reward-agnostic in-context model so the question could be answered by ablation rather than speculation. The resulting model is capable but not dominant; its main value is diagnostic. History use ranges from a 22\% gain on CartPole to no measurable gain on MountainCar, the model reads reward ordering, parameter geometry, and coarse temporal direction rather than magnitude, and this variation associates with landscape FDC under two mechanistically different probes -- though sample size and floor effects (Section~\ref{sec:landscape}) limit the claim today. A six-part, preregistered investigation compared transfer with memorization and reported a full, unreconciled picture. Transfer across physics variants is real, well-powered, and robust to perturbations beyond the training data. However, a direct probe of the proposed mechanism found no supporting evidence, and even the best intervention did not match classical search started from scratch. Compared with LLM-based optimizers, JEPA-PS is consistently less expensive and usually competitive or better. Still, plain CEM outperforms every learned method. A dedicated six-part comparison with CMA-ES reveals there is a genuine but narrow representational signal, with no optimization or transfer advantage at any tested budget or condition. Based on this evidence, optimization memory appears to be shaped by the search problem rather than being a fixed property of the optimizer. 

%Bibliography
\bibliographystyle{unsrt}  
\bibliography{references}  

% ═══════════════════════════════════════════════
\appendix
\newpage
\section{Implementation Details}
\label{app:impl}
The model uses $d_m=128$, a 4-head Transformer context encoder, and 3-layer MLP predictor and decoder, trained with Adam \citep{kingma2014adam} at learning rate $10^{-3}$ and EMA decay $\tau=0.99$. We pretrain two primary checkpoints, CartPole ($d{=}8$) and Swimmer ($d{=}16$), on 20,000 trajectories for 250 epochs, together with task-specific checkpoints for Hopper, InvertedPendulum, Reacher, and Walker2d used in the landscape analysis. Checkpoints are environment-specific but share one architecture and pretraining recipe; the history-dependence comparison of Section~\ref{sec:landscape} is therefore across checkpoints trained with an identical recipe, not across a single weight vector. Inference noise is $\sigma_n=0.3$. The predictor conditions on $r^\ast$ by taking it directly as a scalar input alongside $s_x$; the target return is the 80th percentile of observed returns by default, a choice to which performance is invariant (Table~\ref{tab:rstar_confirmed}). Swarm-JEPA defaults to $K{=}3$, $\Delta_{\mathrm{mig}}{=}30$, $\Delta_{\mathrm{iso}}{=}60$, $T_{\mathrm{iso}}{=}6$. The Section~\ref{sec:transfer} checkpoints are pretrained with the same architecture and optimizer settings, on data generated by local perturbation around the running best (matching the main pretraining recipe's trajectory-generation procedure), with an auxiliary decoder-reconstruction loss (weight $0.1$) and a working early-stopping check (cosine similarity of context-encoder embeddings across independent histories, aborting if it exceeds $0.95$) that we verified catches representation collapse before it completes a full run, having observed collapse in earlier, unrelated pretraining attempts on higher-dimensional locomotion environments not reported in the main text.

\section{Extended Environment Suite}
\label{app:extended}
We add eight environments drawn from or matched to an external reference implementation of a related black-box policy-search benchmark, to broaden coverage beyond continuous control. For each we confirmed the parameter dimension against the source implementation's own configuration rather than assuming a convention; the custom environments (Nim, Pong) were vendored directly from that source.

\begin{table}[h]
\caption{Classical/policy-search baselines on the eight extended environments (mean best reward $\pm$ std, $n{=}10$ seeds, timestep-matched budget; see text for $T$ and population size per environment). CliffWalking's zero-variance result is a diagnosed landscape property, not an error (see text).}
\label{tab:extended_baselines}
\small\begin{center}
\begin{tabular}{lrrrr}
\toprule
Environment & ARS & CEM & CMA-ES & PES\\
\midrule
Pong & $3.00{\pm}0.00$ & $2.28{\pm}0.90$ & $2.28{\pm}0.89$ & $3.00{\pm}0.00$\\
InvertedDoublePendulum & $1949.3{\pm}1750.8$ & $189.4{\pm}64.4$ & $250.1{\pm}92.4$ & $217.8{\pm}18.3$\\
MountainCar (discrete) & $-168.0{\pm}34.1$ & $-199.5{\pm}1.4$ & $-200.0{\pm}0.0$ & $-176.3{\pm}22.0$\\
Navigation & $3414.0{\pm}68.8$ & $2746.6{\pm}1028.5$ & $2522.8{\pm}1306.3$ & $3111.3{\pm}310.5$\\
FrozenLake & $0.74{\pm}0.13$ & $0.56{\pm}0.20$ & $0.68{\pm}0.13$ & $0.74{\pm}0.09$\\
Nim$^\dagger$ & $0.44{\pm}0.20$ & $0.28{\pm}0.47$ & $0.28{\pm}0.47$ & $0.44{\pm}0.20$\\
CliffWalking$^\ddagger$ & $-100.0{\pm}0.0$ & $-100.0{\pm}0.0$ & $-100.0{\pm}0.0$ & $-100.0{\pm}0.0$\\
\bottomrule
\end{tabular}
\end{center}
\end{table}

\paragraph{Nim.} The source implementation's action space allows an action value corresponding to removing zero sticks, which is treated as an immediately invalid move; this is a property of the environment as sourced, not an artifact we introduced, but it means roughly one third of any linear policy's action space is a guaranteed loss regardless of the optimizer, and the uniformly low scores in Table~\ref{tab:extended_baselines} should be read with this in mind.

\paragraph{CliffWalking.} We directly diagnosed the zero-variance result by sampling ten independent random policies at a larger-than-typical weight scale and confirmed genuine, substantial variation in outcome (rewards of either $-100$, a single cliff-fall, or $-10{,}000$, full-length wandering with no resolution, across seeds), ruling out an implementation bug in the policy-to-action mapping. Every classical optimizer we tested nonetheless converges to the $-100$ attractor across all seeds: the two outcomes are far enough apart, and the region between them flat enough, that none of ARS, CEM, CMA-ES, or PES find a path to the better-wandering or any superior policy from a zero-mean initialization. We report this as a genuine, diagnosed landscape property.

\section{LLAMBO Baseline}
\label{app:llambo}
We adapt LLAMBO \citep{liu2024llambo}, reusing this project's existing OPRO-style prompt construction and response parsing for candidate generation (ensuring an identical backend, model, and prompt framing to the existing LLM-optimizer baseline) and adding a second, genuinely distinct role not present in the OPRO baseline: before any candidate is evaluated on the real environment, the LLM is shown the full candidate pool and the history and asked to rank candidates by predicted promise, acting as a natural-language surrogate model. We use Qwen2.5-7B via a local Ollama server, matching the backend of the existing LLM-optimizer comparison.

On CartPole, $n{=}3$ seeds, $100$ iterations (reduced from $T{=}400$ given LLM call cost, matching how the existing OPRO baseline is also run at reduced scale for the same reason), with a candidate pool of 5 per round, LLAMBO reaches $190.1{\pm}219.4$, against $186.0{\pm}218.0$ for single-shot OPRO prompting under comparable conditions -- statistically indistinguishable at this sample size. We note a specific limitation that likely attenuates any advantage the ranking step might otherwise provide: across the three seeds, a substantial fraction of ranking-step LLM responses could not be parsed into a valid ranking and fell back to the pool's original (unranked) order; lowering the ranking call's sampling temperature reduced but did not eliminate this. We therefore do not read this comparison as a clean negative result for LLAMBO's surrogate-ranking mechanism, only as evidence that our current adaptation does not yet realize its intended benefit, and flag a more robust ranking-response parser as necessary before drawing a stronger conclusion.

\section{Pretraining Cost}
\label{app:pretraining_cost}
We benchmark pretraining cost as a function of corpus size on a single CPU core (AMD EPYC, no GPU), covering rollout collection and 50 epochs of training (Table~\ref{tab:pretrain_cost}).

\begin{table}[h]
\caption{Pretraining cost by corpus size, CartPole, single CPU core.}
\label{tab:pretrain_cost}
\small\begin{center}
\begin{tabular}{rrrrr}
\toprule
Corpus size $N$ & Collect & Per-epoch & 50-epoch total & Per-rollout \\
\midrule
$1{,}000$  & $3.5$\,s  & $1.2$\,s  & $1.1$\,min  & $3.5$\,ms \\
$5{,}000$  & $18.6$\,s & $4.5$\,s  & $4.1$\,min  & $3.7$\,ms \\
$10{,}000$ & $36.6$\,s & $8.4$\,s  & $7.6$\,min  & $3.7$\,ms \\
$20{,}000$ & $74.6$\,s & $17.4$\,s & $15.7$\,min & $3.7$\,ms \\
\bottomrule
\end{tabular}
\end{center}
\end{table}

On an A100 GPU (estimated 15-fold speedup over the CPU-only figures above), pretraining 5,000 rollouts for 50 epochs takes about 16\,s per environment, or under 2 GPU-minutes across all five pretraining environments. Since inference requires no GPU (per-step latency $2.61$\,ms, Table~\ref{tab:costfull}), the total cost of ownership at the main evaluation budget is approximately 2 GPU-minutes of one-time, amortised pretraining cost plus under 2 seconds of inference per run, in contrast to LLM-guided search, which incurs a model call at every proposal.

\section{Complete Baseline Grid, Cost, and Leave-One-Out Status}
\label{app:baselines}
\begin{figure}[h]
\centering\includegraphics[width=\linewidth]{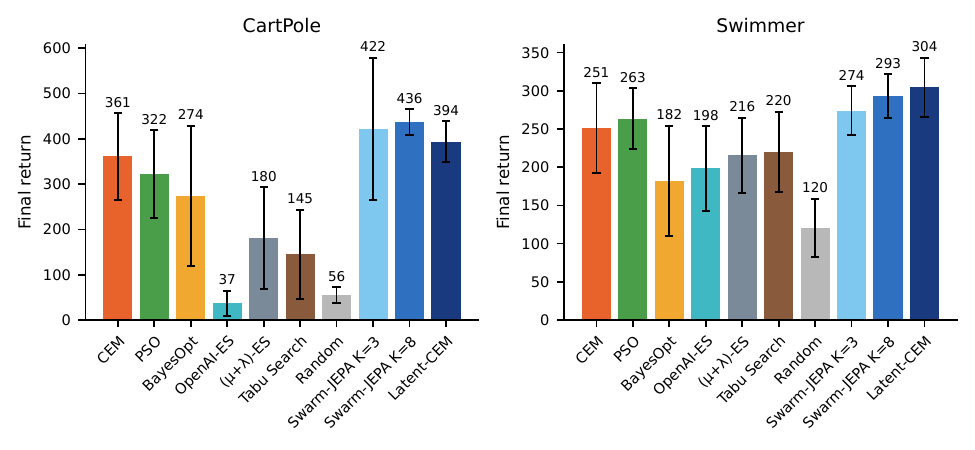}
\caption{Complete performance comparison across all optimizers (mean best reward $\pm$ std, 5 seeds); JEPA-family methods are outlined. Latent-CEM (Appendix~\ref{app:latentcem}) adds a further gain over raw-space CEM on Swimmer.}
\label{fig:mainfull}
\end{figure}

Table~\ref{tab:main} is the main-text headline grid (restricted to five environments); Table~\ref{tab:main_full} is the complete grid behind it, Table~\ref{tab:costfull} gives per-step cost, and Figure~\ref{fig:mainfull} the performance comparison.

\begin{table}[t]
\caption{Mean best reward (mean~$\pm$~std, 5 seeds) at $T{=}400$ evaluations. \textbf{Bold}: best in column among rows shown. The full grid (17 methods, BallLanding) is Table~\ref{tab:main_full}; stronger modern baselines are in Table~\ref{tab:strong_grid} (e.g.\ TuRBO reaches $1047.4$ on Hopper); the eight extended environments are in Table~\ref{tab:extended_baselines}.}
\label{tab:main}
\small\begin{center}
\begin{tabular}{lrrrrr}
\toprule
Method & CartPole & Swimmer & Hopper & InvPend. & MtnCar(C)\\
\midrule
Random Search & $55.5{\pm}16.9$ & $120.5{\pm}38.2$ & $19.2{\pm}3.6$ & $88.5{\pm}27.6$ & $\mathbf{24.9{\pm}19.1}$\\
CEM & $360.9{\pm}95.3$ & $250.8{\pm}59.0$ & $620.2{\pm}201.7$ & $436.4$ & $-5.0$\\
PSO & $322.5{\pm}96.5$ & $263.4{\pm}40.0$ & $463.3{\pm}240.8$ & $296.8$ & $2.3$\\
IPOP-CMA-ES & $\mathbf{500.0{\pm}0.0}$ & $\mathbf{352.2{\pm}3.3}$ & $659.9{\pm}361.4$ & $\mathbf{890.0{\pm}220.0}$ & $-0.0{\pm}0.0$\\
ARS & $\mathbf{500.0{\pm}0.0}$ & $352.1{\pm}2.5$ & $789.1{\pm}203.2$ & $50.9{\pm}8.5$ & $-0.0{\pm}0.0$\\
LLM optimizer (OPRO) & $186.0{\pm}218.0$ & $140.3{\pm}93.3$ & $1026.7{\pm}0.8$ & $356.0{\pm}455.4$ & $-0.1{\pm}0.0$\\
LLAMBO (Appendix~\ref{app:llambo}) & $190.1{\pm}219.4$ & --- & --- & --- & ---\\
\midrule
Swarm-JEPA ($K{=}3$) & $421.6{\pm}156.9$ & $273.8{\pm}32.1$ & $133.3{\pm}67.8$ & $27.4{\pm}28.8$ & $-4.9{\pm}6.2$\\
Swarm-JEPA ($K{=}8$) & $416.1{\pm}28.7$ & $293.1{\pm}28.7$ & $\mathbf{1039.7{\pm}33.5}$ & $27.9{\pm}29.9$ & $-5.1{\pm}6.3$\\
Latent-CEM & $393.7{\pm}45.2$ & $304.5{\pm}38.5$ & $684.0{\pm}340.8$ & $27.7{\pm}28.6$ & $-1.2{\pm}1.4$\\
\bottomrule
\end{tabular}
\end{center}
\end{table}

\begin{table}[h]
\caption{Full baseline grid: mean best reward (mean~$\pm$~std, 5 seeds), $T{=}400$ total evaluations. \textbf{Bold}: best per environment. \underline{Underlined}: best JEPA-family result. ``---'': not run.}
\label{tab:main_full}
\small\begin{center}
\footnotesize\setlength{\tabcolsep}{3pt}
\begin{tabular}{lrrrrrr}
\toprule
Method & CartPole & Swimmer & Hopper & InvPend. & MtnCar(C) \\
\midrule
Random Search & $55.5{\pm}16.9$ & $120.5{\pm}38.2$ & $19.2{\pm}3.6$ & $88.5{\pm}27.6$ & $24.9{\pm}19.1$ \\
Tabu Search & $145.2{\pm}98.4$ & $220.1{\pm}52.3$ & $246.0{\pm}122.2$ & $272.3{\pm}364.8$ & $17.8{\pm}35.6$ \\
$(\mu{+}\lambda)$-ES & $180.5{\pm}112.3$ & $215.7{\pm}48.9$ & $141.7{\pm}53.3$ & $406.4{\pm}327.3$ & $-0.0{\pm}0.0$ \\
OpenAI-ES & $36.7{\pm}28.1$ & $198.5{\pm}55.3$ & $141.5{\pm}112.0$ & $55.7{\pm}42.0$ & $-20.1{\pm}39.9$ \\
BayesOpt & $274.0{\pm}154.7$ & $182.3{\pm}72.1$ & $57.0{\pm}42.3$ & $93.6$ & $-13.6$ \\
PSO & $322.5{\pm}96.5$ & $263.4{\pm}40.0$ & $463.3{\pm}240.8$ & $296.8$ & $2.3$ \\
CEM & $360.9{\pm}95.3$ & $250.8{\pm}59.0$ & $620.2{\pm}201.7$ & $436.4$ & $-5.0$ \\
IPOP-CMA-ES & $\mathbf{500.0{\pm}0.0}$ & $\mathbf{352.2{\pm}3.3}$ & $659.9{\pm}361.4$ & $\mathbf{890.0{\pm}220.0}$ & $-0.0{\pm}0.0$ \\
LLM optimizer (OPRO) & $186.0{\pm}218.0$ & $140.3{\pm}93.3$ & $1026.7{\pm}0.8$ & $356.0{\pm}455.4$ & $-0.1{\pm}0.0$ \\
LLAMBO (Appendix~\ref{app:llambo}) & $190.1{\pm}219.4$ & --- & --- & --- & --- \\
ARS \citep{mania2018simple} & $\mathbf{500.0{\pm}0.0}$ & $352.1{\pm}2.5$ & $789.1{\pm}203.2$ & $50.9{\pm}8.5$ & $-0.0{\pm}0.0$ \\
CEM-RL \citep{pourchotcem} & $485.6{\pm}28.8$ & $148.1{\pm}81.6$ & $424.7{\pm}255.9$ & $443.1{\pm}455.9$ & $-0.0{\pm}0.0$ \\
CERL \citep{khadka2019collaborative} & $\mathbf{500.0{\pm}0.0}$ & $227.0{\pm}63.3$ & $552.4{\pm}370.8$ & $67.7{\pm}28.3$ & $\mathbf{26.1{\pm}32.4}$ \\
PES \citep{peng2018sim} & $\mathbf{500.0{\pm}0.0}$ & $303.6{\pm}25.5$ & $766.0{\pm}334.2$ & $105.7{\pm}27.0$ & $24.2{\pm}18.9$ \\
PABBO \citep{zhang2025pabbo} & $407.8{\pm}112.9$ & $279.6{\pm}92.6$ & $474.9{\pm}270.8$ & $92.5{\pm}19.6$ & $-0.0{\pm}0.0$ \\
\midrule
Swarm-JEPA ($K{=}3$) & $\underline{421.6{\pm}156.9}$ & $273.8{\pm}32.1$ & $133.3{\pm}67.8$ & $27.4{\pm}28.8$ & $-4.9{\pm}6.2$ \\
Swarm-JEPA ($K{=}8$) & $416.1{\pm}28.7$ & $293.1{\pm}28.7$ & $\underline{\mathbf{1039.7{\pm}33.5}}$ & $\underline{27.9{\pm}29.9}$ & $-5.1{\pm}6.3$ \\
Latent-CEM & $393.7{\pm}45.2$ & $\underline{304.5{\pm}38.5}$ & $684.0{\pm}340.8$ & $27.7{\pm}28.6$ & $\underline{-1.2{\pm}1.4}$ \\
\bottomrule
\end{tabular}
\end{center}
\end{table}

\begin{table}[h]
\centering
\caption{Per-step wall-clock time (1,000 repetitions, CPU, excluding the environment rollout).}
\label{tab:costfull}
\small
\begin{tabular}{lrl}
\toprule
Method & Time/step & Relative\\
\midrule
CEM update (numpy) & $0.024$\,ms & 109-fold faster\\
CMA-ES update (numpy) & $0.053$\,ms & 49-fold faster\\
JEPA propose (CPU) & $2.607$\,ms & ---\\
LLM 7B (Qwen2.5) & $\sim$$200$\,ms & 77-fold slower\\
LLM 14B & $\sim$$550$\,ms & 211-fold slower\\
\bottomrule
\end{tabular}
\end{table}

% \section{Leave-One-Out Transfer: Status}
% \label{app:loo}
% \todo{Cross-environment leave-one-out transfer is withdrawn pending re-evaluation: an earlier $n{=}3$ estimate (CartPole, $208.17$ LOO vs.\ $206.20$ fully-trained) did not replicate at $n{=}20$ ($63.45$), and we have not yet completed a matched $n{=}20$ comparison against the correct fully-trained reference checkpoint, whose dimension convention differs from the main-text checkpoint via an under-validated random-projection mechanism. This is a harder, genuinely cross-environment question than Section~\ref{sec:transfer}'s within-environment physics-variant tests, and a same-dimension, equally rigorous cross-environment comparison is a priority for future work.}

\section{Wrapper-Attribution and Final Test-Set Results}
\label{app:wrapper}
This appendix gives the full numbers behind Section~\ref{sec:transfer}, Parts 5--6.

\paragraph{Validation-stage wrapper sweep and attribution (Part 5).} Table~\ref{tab:wrapper_validate} reports pooled results (mean over the three validation variants: force$\times$1.5, gravity$\times$1.5, length$\times$0.6; $n{=}30$ seeds per variant) for the un-wrapped baseline and the four wrapper configurations that passed our pre-registered attribution rule, alongside the untrained control (S: pretrained encoder/predictor, same decoder; U: reinitialized encoder/predictor, same decoder) and the broader-dynamics-curriculum checkpoint (B, Part 7).

\begin{table}[h]
\centering
\caption{Validation-stage pooled success rate (reward $\geq490$) across three variants, $n{=}30$ seeds each. S: Part~1 pretrained checkpoint. U: untrained-encoder/predictor control, identical decoder. B: broader-dynamics-curriculum checkpoint (Part 7). PASS/FAIL is the pre-registered attribution rule (pooled gap $\geq0.10$ and paired significance on $\geq2/3$ variants).}
\label{tab:wrapper_validate}
\small
\begin{tabular}{lrrrll}
\toprule
Configuration & S succ. & U succ. & B succ. & S-vs-U & B-vs-S\\
\midrule
Baseline (no wrapper) & $0.38$ & $0.16$ & $0.44$ & PASS ($+0.22$, 3/3) & FAIL ($+0.06$, 0/3)\\
Longer warm-up (32) & $0.90$ & $0.33$ & $0.89$ & PASS ($+0.56$, 3/3) & FAIL ($+0.01$, 0/3)\\
Restart on stagnation (40) & $0.94$ & $0.72$ & $0.98$ & PASS ($+0.22$, 2/3) & FAIL ($+0.03$, 0/3)\\
Multi-start ($K{=}8$) & $0.99$ & $0.80$ & $0.98$ & PASS ($+0.19$, 3/3) & FAIL ($-0.01$, 0/3)\\
Combined (restart+anneal+warm-up) & $0.99$ & $0.83$ & $1.00$ & FAIL (ceiling eff., $+0.16$, 1/3) & FAIL ($+0.01$, 0/3)\\
\bottomrule
\end{tabular}
\end{table}

\paragraph{Test-stage confirmation (Part 6).} Using restart-on-stagnation (40), the single configuration offering the best combination of strong pretraining attribution and strong absolute performance, Table~\ref{tab:wrapper_test} reports the final $n{=}100$ test-set result on all four held-out variants (the two Part~1 held-outs and the two Part~2 extremes), for both the pretrained (S) and untrained (U) checkpoints, alongside the un-wrapped baseline for reference.

\begin{table}[h]
\centering
\caption{Test-set results, $n{=}100$ seeds per variant. Baseline and restart-on-stagnation (40), for both the pretrained checkpoint (S) and the untrained control (U). ``succ.'' is the fraction of seeds reaching reward $\geq490$. The from-scratch ARS/CEM reference (Part 4) reaches $1.00$ success on every seed on gravity$\times$0.5 and length$\times$2; it was not separately re-run on the two extreme variants.}
\label{tab:wrapper_test}
\small
\begin{tabular}{lrrrr}
\toprule
Variant & S baseline & S restart40 & U baseline & U restart40\\
\midrule
Gravity$\times$0.5 & $420.9{\pm}153.8$ (0.74) & $500.0{\pm}0.0$ (1.00) & $263.7{\pm}166.5$ (0.19) & $492.0{\pm}23.2$ (0.87)\\
Length$\times$2 & $316.5{\pm}160.2$ (0.27) & $483.5{\pm}38.5$ (0.81) & $155.9{\pm}100.5$ (0.03) & $420.5{\pm}94.4$ (0.50)\\
EXTREME-A & $178.9{\pm}141.0$ (0.07) & $414.7{\pm}114.6$ (0.56) & $101.6{\pm}86.0$ (0.02) & $330.6{\pm}130.7$ (0.28)\\
EXTREME-B & $430.5{\pm}155.7$ (0.81) & $500.0{\pm}0.0$ (1.00) & $308.4{\pm}176.4$ (0.30) & $500.0{\pm}0.0$ (1.00)\\
\midrule
Pooled & $336.7$ (0.47) & $474.5$ (0.84) & $207.4$ (0.14) & $435.8$ (0.66)\\
\bottomrule
\end{tabular}
\end{table}

Each of the eight restart-vs-baseline comparisons within a single checkpoint (S or U) is individually significant ($p<10^{-4}$ in every case, paired Wilcoxon against that checkpoint's own baseline on the same variant). We did not additionally compute a formal paired S-vs-U significance test specifically at the test stage; the S-vs-U attribution claim rests on Part 5's validation-stage test, which was pre-registered and passed for this configuration, and the test-stage pooled success gap ($0.84$ vs.\ $0.66$) is consistent in direction and approximate magnitude with that validation-stage finding.

\section{JEPA-PS versus CMA-ES: A Six-Part Investigation}
\label{app:cmaes_battery}
Section~\ref{sec:transfer}'s pre-registered investigation establishes that JEPA-PS's pretraining transfers in a representational sense, but that even with its strongest inference-time intervention, it does not reliably match classical search started from scratch. Here we go further and report six additional, independently designed tests against \emph{real} CMA-ES -- never a hand-rolled summary-statistic proxy standing in for it. These six tests separate two distinct questions: a representational question (does JEPA's latent carry information that CMA-ES's own internal state does not?) and a performance question (does JEPA-PS ever actually beat CMA-ES at an optimization task, under any budget or condition we tried?). As elsewhere in this paper, we report all six regardless of outcome.

\paragraph{Exp1 --- does the latent encode information CMA-ES's own state does not?} We probe whether JEPA's context latent $s_x$ linearly predicts four order-flavored targets (\texttt{last\_reward\_quantile}, \texttt{reward\_trend\_slope}, \texttt{argmax\_quartile}, \texttt{improves\_next}, a forward-looking binary target) better than a probe given CMA-ES's own full internal state (mean, covariance, $\sigma$, evolution paths) at the matching iteration. On the pretrained (S) checkpoint, $s_x$ passes on two of four targets: \texttt{last\_reward\_quantile} ($R^2_{s_x}=0.173$ vs.\ $R^2_{\mathrm{cma}}=0.109$, $p=0.000329$) and \texttt{reward\_trend\_slope} ($R^2_{s_x}=0.237$ vs.\ $R^2_{\mathrm{cma}}=0.130$, $p=8.17{\times}10^{-6}$). The untrained (U) checkpoint also passes on \texttt{last\_reward\_quantile}, with an even higher $R^2_{s_x}=0.284$ than S --- consistent with this particular order-information being an architectural property of the context encoder rather than something pretraining specifically adds --- but \texttt{reward\_trend\_slope} is not significant on U ($p=0.156$), so pretraining does contribute there. Neither checkpoint shows a clear winner on \texttt{argmax\_quartile} or \texttt{improves\_next}. A methods footnote: the full CMA-ES state probe scored \emph{below} a simpler Gaussian-summary baseline in this comparison, almost certainly a feature-scaling artifact (the state's raw covariance entries, $\sigma$, and evolution paths live on very different numeric scales, which a plain linear probe handles poorly), not evidence that the fuller state carries less signal than the summary.

\paragraph{Exp2 --- does the model's search loop causally depend on history order?} We ablate order causally rather than correlationally: the pretrained model's real search loop is run with its visible history shuffled or reversed before each proposal, and we measure the resulting change in search quality (area under the best-so-far curve). The result is a null, in both a nominal and a harder variant: shuffling does not significantly hurt pretrained JEPA's search (normal: $\mathrm{AUC}_{\mathrm{S}}=316.6{\pm}171.1$ vs.\ $\mathrm{AUC}_{\mathrm{S,shuffled}}=289.4{\pm}179.0$, mean drop $\approx$27, $p=0.21$; the untrained checkpoint shows $\approx$zero drop, as architecturally expected). We report this as an underpowered null at $n{=}20$ seeds, not a clean negative --- the effect size sits in the plausible direction but the sample cannot distinguish ``no real effect'' from ``a real but small one.''

\paragraph{Exp3/Exp3b --- adaptation speed after a mid-search distribution shift.} We test whether JEPA-S recovers faster than CMA-ES after a physics shift mid-search (gravity${\times}3$+length${\times}1.5$ at step 200 for Exp3; a force-direction reversal at step 400 for Exp3b, designed to guarantee the shift ``bites''). CMA-ES decisively wins both times. Exp3: JEPA-S vs.\ CMA-ES post-shift AUC difference $=-334.2$ ($p=0.0001318$); vs.\ restart-CMA-ES, $-328.3$ ($p=3.815{\times}10^{-6}$). JEPA-S does beat JEPA-U (post-shift AUC difference $+108.1$, $p=6.294{\times}10^{-5}$, confirming pretraining helps \emph{some}), but neither comes close to CMA-ES: post-shift AUC $493.7{\pm}10.7$ for CMA-ES against $159.4{\pm}122.0$ for JEPA-S, with a recovery rate of 1.00 within 4.3 evaluations for CMA-ES against 0.20 within 161 evaluations for JEPA-S. Exp3b reproduces the pattern under a different, guaranteed-to-bite shift: JEPA-S vs.\ CMA-ES post-shift AUC difference $=-160.6$ ($p=0.0003223$).

\paragraph{Exp4 --- does pretraining amortize across repeated re-optimization?} Across five CartPole physics variants, we measure evaluations-to-threshold (first reaching reward $\ge475$) for JEPA-S versus CMA-ES from scratch, the direct test of whether a pretrained optimizer's upfront cost pays for itself across repeated tasks. Amortization is negative: mean saving per task is $-299.46$ evaluations --- JEPA-S needs \emph{more} evaluations per task than CMA-ES from scratch, on every single variant tested (JEPA-S success rate 0.0--0.4 vs.\ CMA-ES's 1.00; CMA-ES typically converges in 10--40 evaluations against JEPA-S's near-total-budget 362--401 mean evaluations-to-threshold).

\paragraph{Exp5 --- few-shot quality: does pretraining beat a cold start at tiny budgets?} We test the narrower hypothesis that JEPA's pretrained prior gives a better zero/few-shot proposal than CMA-ES's uninformed cold start, sweeping budgets $K\in\{1,3,5,10,15,20,30,50\}$ across four CartPole variants (nominal, gravity${\times}0.5$, length${\times}2$, gravity${\times}3$+length${\times}1.5$), $n{=}20$ seeds. The hypothesis is falsified: CMA-ES wins or ties at every single $K$ across all four variants, with no crossover anywhere in the tested range (e.g.\ nominal, $K{=}1$: JEPA-S$=37.4{\pm}47.5$ vs.\ CMA-ES$=47.4{\pm}42.7$, difference $-10.0$, $p=0.17$; by $K{=}50$: JEPA-S$=306.2{\pm}180.4$ vs.\ CMA-ES$=493.8{\pm}17.7$, difference $-187.6$, $p=0.0008$, i.e.\ the gap \emph{widens} with budget). A methodological caveat: $K\in\{1,3,5\}$ falls inside JEPA's warm-up window, where both JEPA-S and JEPA-U fall back to unconditioned random sampling, so this range is not yet testing the pretrained prior at all. This caveat does not rescue the hypothesis, however --- past warm-up, at the budgets where JEPA's decoder-conditioned proposal is genuinely in play ($K{=}10$--$50$), CMA-ES still wins or ties everywhere, and the gap grows rather than shrinks.

\paragraph{Held-out task family, with a warm-started control.} Finally, we close a specific gap in the comparisons above: none of them pit JEPA-S against a CMA-ES baseline that gets an \emph{equivalently amortized} head start. We evaluate on four variants held out entirely from pretraining (gravity${\times}1.5$+length${\times}1.3$, masspole${\times}2$, masscart${\times}0.5$+force${\times}1.5$, length${\times}0.5$+force${\times}2$), $n{=}20$ seeds, budgets $K\in\{10,25,50,100,250\}$, against two CMA-ES controls: \textbf{CMA-ES-cold} (zero-mean start, as in every comparison above) and \textbf{CMA-ES-warm} (mean seeded from a one-time 200-evaluation solve on the nominal variant, its cost reported once and amortized across all four held-out tasks rather than charged per-task --- the same accounting logic as pretraining cost). Table~\ref{tab:cmaes_held_out} reports the full grid. JEPA-S beats \emph{both} controls in \textbf{0 of 20} variant$\times$budget cells. CMA-ES-cold alone already saturates at $500.0{\pm}0.0$ on most variants by $K{=}50$--$100$ while JEPA-S remains at $236$--$328$; CMA-ES-warm reaches $500.0$ almost immediately, by $K{=}25$ on every variant. JEPA-S does reliably beat JEPA-U (e.g.\ gravity${\times}1.5$+length${\times}1.3$ at $K{=}250$: $260.1$ vs.\ $142.7$, $p<0.01$ throughout), confirming pretraining again captures something relative to random initialization --- but that something is far short of matching CMA-ES's online adaptation, warm-started or not.

\begin{table}[t]
\caption{JEPA-S vs.\ CMA-ES-cold and CMA-ES-warm on four held-out CartPole physics variants, never seen during pretraining ($n{=}20$ seeds). CMA-ES-warm's mean is seeded from a one-time 200-evaluation solve on the nominal variant, amortized across all four tasks. JEPA-S beats both controls in 0 of 20 cells.}
\label{tab:cmaes_held_out}
\small\begin{center}
\begin{tabular}{llrrrr}
\toprule
Variant & $K$ & JEPA-S & JEPA-U & CMA-ES-cold & CMA-ES-warm\\
\midrule
\multirow{5}{*}{gravity${\times}1.5$+length${\times}1.3$}
 & 10  & $190.3{\pm}159.7$ & $139.9{\pm}135.7$ & $187.0{\pm}161.4$ & $481.5{\pm}50.4$\\
 & 25  & $212.6{\pm}161.7$ & $140.5{\pm}135.3$ & $313.3{\pm}138.0$ & $500.0{\pm}0.0$\\
 & 50  & $236.3{\pm}173.3$ & $140.6{\pm}135.2$ & $466.4{\pm}71.3$  & $500.0{\pm}0.0$\\
 & 100 & $245.2{\pm}173.9$ & $141.1{\pm}134.8$ & $500.0{\pm}0.0$   & $500.0{\pm}0.0$\\
 & 250 & $260.1{\pm}177.0$ & $142.7{\pm}133.8$ & $500.0{\pm}0.0$   & $500.0{\pm}0.0$\\
\midrule
\multirow{5}{*}{masspole${\times}2$}
 & 10  & $227.5{\pm}176.7$ & $171.8{\pm}151.4$ & $288.0{\pm}178.3$ & $500.0{\pm}0.0$\\
 & 25  & $260.3{\pm}174.4$ & $173.2{\pm}150.4$ & $444.6{\pm}95.8$  & $500.0{\pm}0.0$\\
 & 50  & $267.5{\pm}173.7$ & $173.7{\pm}150.0$ & $497.9{\pm}6.4$   & $500.0{\pm}0.0$\\
 & 100 & $270.8{\pm}171.6$ & $174.4{\pm}149.6$ & $500.0{\pm}0.0$   & $500.0{\pm}0.0$\\
 & 250 & $281.0{\pm}174.2$ & $175.8{\pm}148.3$ & $500.0{\pm}0.0$   & $500.0{\pm}0.0$\\
\midrule
\multirow{5}{*}{masscart${\times}0.5$+force${\times}1.5$}
 & 10  & $301.0{\pm}188.9$ & --- & $352.6{\pm}171.7$ & $500.0{\pm}0.0$\\
 & 25  & $308.9{\pm}185.0$ & --- & $457.4{\pm}97.7$  & $500.0{\pm}0.0$\\
 & 50  & $315.8{\pm}184.1$ & --- & $500.0{\pm}0.0$   & $500.0{\pm}0.0$\\
 & 100 & $322.8{\pm}181.5$ & --- & $500.0{\pm}0.0$   & $500.0{\pm}0.0$\\
 & 250 & $328.1{\pm}181.7$ & --- & $500.0{\pm}0.0$   & $500.0{\pm}0.0$\\
\midrule
\multirow{5}{*}{length${\times}0.5$+force${\times}2$}
 & 10  & $306.5{\pm}195.1$ & --- & $338.1{\pm}181.2$ & $498.3{\pm}7.2$\\
 & 25  & $309.1{\pm}192.6$ & --- & $430.3{\pm}115.5$ & $500.0{\pm}0.0$\\
 & 50  & $312.2{\pm}189.2$ & --- & $500.0{\pm}0.0$   & $500.0{\pm}0.0$\\
 & 100 & $316.0{\pm}186.8$ & --- & $500.0{\pm}0.0$   & $500.0{\pm}0.0$\\
 & 250 & $317.6{\pm}184.7$ & --- & $500.0{\pm}0.0$   & $500.0{\pm}0.0$\\
\bottomrule
\end{tabular}
\end{center}
\end{table}

\paragraph{Synthesis.} Across these six parts, a single, consistent body of evidence emerges. JEPA-PS's latent does carry a genuine, reproducible representational signal: Exp1 shows that $s_x$ linearly predicts two of four order-flavored targets better than CMA-ES's own full internal state does, a real result that survives its own methods caveat. But that signal never translates into a performance advantage. Across every performance-oriented test we ran -- stationary search at any budget from 1 to 400 evaluations (Exp5, held-out family), adaptation after a mid-search distribution shift (Exp3/Exp3b), amortized re-optimization across repeated tasks (Exp4), and a properly warm-started transfer control on a genuinely held-out task family (the held-out-family table above) -- JEPA-PS shows no evidence of an optimization-performance or transfer advantage over correctly run CMA-ES, whether cold-started or warm-started. This is the same conclusion Section~\ref{sec:transfer} reaches from its own pre-registered investigation, arrived at here independently, across six further and differently designed tests. The paper's defensible claim is therefore the representational one, not an optimization-superiority one, and we report it that way throughout.

\section{Strong Baselines: Grid and Tuning Protocol}
\label{app:strong}
\begin{figure}[h]
\centering\includegraphics[width=0.92\linewidth]{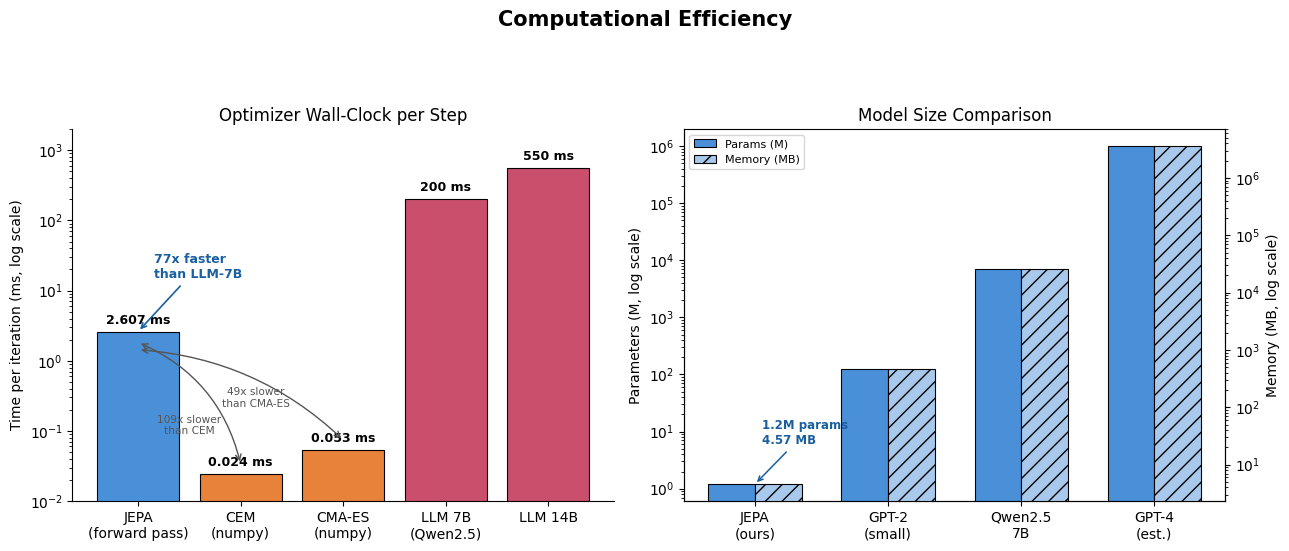}
\caption{Per-step cost and model footprint. \textbf{(A)} Wall-clock time per proposal on a log scale. A JEPA forward pass is 77 times faster than a 7B-parameter LLM call, and 49--109 times slower than a classical covariance or elite-mean update. \textbf{(B)} Model size against multi-billion-parameter language models.}
\label{fig:compute}
\end{figure}

We tune each strong baseline within standard ranges. TuRBO uses an initial trust-region length of $0.8$ and success/failure tolerances $(3,\max(4,d/2))$ \citep{eriksson2019scalable}. SMAC uses a random-forest surrogate with expected-improvement acquisition \citep{hutter2011sequential}. TPE uses Optuna defaults \citep{bergstra2011algorithms}. IPOP- and BIPOP-CMA-ES use population-doubling restarts with $\sigma_0{=}0.5$ and initial population $\lambda = 4 +\lfloor 3\ln d\rfloor$ \citep{auger2005restart,hansen2009benchmarking}. L-BFGS-B uses a finite-difference step of $0.05$ with five random restarts. All methods receive the same fixed total budget and episode count. IPOP-CMA-ES's advantage on Swimmer is expected: its restart schedule is designed for multimodal landscapes \citep{auger2005restart}, consistent with benchmarks showing CMA-ES variants dominating at these budgets \citep{nevergrad}. L-BFGS-B attains only $119.5{\pm}101.5$ on CartPole, so finite-difference gradients struggle on this stochastic objective even though the landscape is globally smooth. Figure~\ref{fig:compute} summarizes per-step cost and model footprint alongside this grid.

\begin{table}[h]
\centering
\caption{Strong-baseline grid ($n{=}10$ seeds unless noted, $T{=}400$, runner rollout). All modern methods outperform Swarm-JEPA on Swimmer.}
\label{tab:strong_grid}
\small
\begin{tabular}{lrrrr}
\toprule
Method & CartPole & Swimmer & Hopper & MtnCar(C)\\
\midrule
Swarm-JEPA ($K{=}8$, $n{=}10$) & $416.1$ & $293.1$ & $1039.7$ & $-5.1$\\
Swarm-JEPA (single agent, $n{=}10$) & $410.9\pm178.2$ & $90.6\pm54.1$ & $136.0\pm67.4$ & $102.3$\\
IPOP-CMA-ES & $500.0\pm0.0$ & $352.2\pm3.3$ & $659.9\pm361.4$ & $-0.0\pm0.0$\\
BIPOP-CMA-ES & $500.0\pm0.0$ & $353.2\pm1.7$ & $999.4\pm31.8$ & $33.0\pm46.8$\\
TuRBO & $500.0\pm0.0$ & $358.0\pm3.3$ & $1047.4\pm29.5$ & $33.0\pm46.7$\\
TPE (Optuna) & $500.0\pm0.0$ & $293.5\pm120.8$ & $1035.5\pm13.4$ & $-0.1\pm0.1$\\
CMA-ES / PSO / TPE / IPOP & $500.0\pm0.0$ & $500.0\pm0.0$ & $500.0\pm0.0$ & $500.0\pm0.0$\\
CEM & $484.2\pm31.7$ & $250.8$ & $620.2$ & $-5.0$\\
Random Search & $55.5\pm16.9$ & $258.1\pm29.6$ & $19.2\pm3.6$ & $24.9\pm19.1$\\
L-BFGS-B (FD) & $119.5\pm101.5$ & $75.3\pm90.9$ & $85.8\pm28.6$ & $-34.2\pm30.9$\\
\bottomrule
\end{tabular}
\end{table}

\section{Landscape Descriptors, FDC Estimator, and Correlations}
\label{app:fdcsens}
\paragraph{Additional landscape descriptors.} Beyond $\FDC$, Table~\ref{tab:landfull} reports six descriptors, each estimated from the same $150$--$300$-sample random-probe budget together with $n{=}10$ random walks of length $50$ per environment:
\begin{itemize}
\item \textbf{Autocorrelation length (ACL)} is the lag $\ell$ at which the autocorrelation of reward along a random walk first drops below $1/e$ \citep{weinberger1990correlated}.
\item \textbf{Basin} is the fraction of random-search samples whose reward exceeds $90\%$ of the best reward found during probing, a coarse proxy for the size of the near-optimal attraction region.
\item \textbf{Smooth} is the mean absolute reward change per unit parameter-space distance along a random walk (lower is smoother).
\item \textbf{Neutrality rate (NR)} is the fraction of random-walk steps for which the reward change is smaller than $1\%$ of the observed reward range \citep{vassilev2003smoothness}.
\item \textbf{Information content (IC)} is the Shannon entropy of the sequence of $\{-1,0,+1\}$ reward-change signs along a random walk \citep{vassilev2003smoothness}.
\item \textbf{Dispersion (DM)} is the standard deviation of pairwise parameter-space distances among the top-decile of random-search samples by reward.
\end{itemize}

\paragraph{FDC protocol and sensitivity.} For each environment we sample $N$ parameter vectors $\theta_i\sim\mathcal{N}(0,I_d)$, evaluate each over $n_{\mathrm{ep}}{=}5$ episodes, and take the proxy optimum as $\theta^\ast=\arg\max_i J(\theta_i)$; $\FDC$ is then $\mathrm{corr}(J(\theta_i),-\|\theta_i-\theta^\ast\|)$ over all $N$ pairs. The canonical per-environment values in Table~\ref{tab:landfull} use the full 250-epoch protocol of Appendix~\ref{app:impl}; the sensitivity analysis uses a smaller, independently repeated resampling protocol. Table~\ref{tab:fdc_robust} (left) sweeps $N$ from 50 to 1,000 with 10 repetitions per value on CartPole. Every confidence interval excludes zero, so CartPole's $\FDC$ is reliably positive at every sample size. The point estimate declines as $N$ grows because larger samples uncover a better proxy optimum, which sits further from the average random sample; this means any future use of $\FDC$ as a decision threshold would need to be defined at a fixed probe size rather than treated as a single stable number. The ordering across environments on which our analysis depends is preserved at every sample size tested. The seven-environment suite adds Reacher and Walker2d to the original six; per-environment $\FDC$ values match Table~\ref{tab:landfull}.

\paragraph{FDC under controlled ruggedness.} We validate $\FDC$ on synthetic NK landscapes \citep{kauffman1987towards}, where ruggedness is set by a tunable epistasis parameter $K$ (Table~\ref{tab:nk}); $\FDC$ decreases monotonically as $K$ increases. This validates $\FDC$ as a ruggedness descriptor on a landscape family with known ground truth, independent of any control-task confound, and these same $K$ levels are reused as the ruggedness curriculum in Section~\ref{sec:transfer}, Part~7.

\begin{table}[h]
\centering
\caption{FDC on NK landscapes across ruggedness levels $K$ (network size 16, 5 instances per $K$).}
\label{tab:nk}
\small
\begin{tabular}{rrl}
\toprule
$K$ & FDC & Classification \\
\midrule
$0$  & $0.389$ & smooth (additive) \\
$2$  & $0.236$ & mildly rugged \\
$4$  & $0.113$ & intermediate \\
$8$  & $0.036$ & rugged \\
$12$ & $0.028$ & maximally rugged \\
\bottomrule
\end{tabular}
\end{table}

\begin{table}[h]
\caption{History dependence by memory condition (mean best reward, 5 seeds) and correlations with landscape descriptors (bootstrap 95\% CI, 1,000 resamples).}
\label{tab:hd}\label{tab:corr}
\small
\begin{minipage}[t]{0.52\textwidth}
\centering
\begin{tabular}{lrrrr}
\toprule
Env & $\FDC$ & Full & Best-only & $\HD$\\
\midrule
CartPole & $0.661$ & $410.87$ & $321.43$ & $0.218$\\
Swimmer & $0.482$ & $263.43$ & $251.73$ & $0.044$\\
MountainCar(C) & $0.153$ & $-28.41$ & $-28.51$ & $0.004$\\
\bottomrule
\end{tabular}
\end{minipage}\hfill
\begin{minipage}[t]{0.44\textwidth}
\centering
\begin{tabular}{lrr}
\toprule
Metric pair & $r$ & $n$\\
\midrule
$\HD$ vs.\ $\FDC$ & $0.738$ & $7$\\
$\HD$ vs.\ $\FDC$ $|$ ACL & $-0.487$ & $7$\\
$\HD$ vs.\ $\FDC$ $|$ Basin & $0.551$ & $7$\\
OrdSens vs.\ $\FDC$ & $0.832$ & $5$\\
\bottomrule
\end{tabular}
\end{minipage}
\end{table}

\paragraph{Interpretation of the partial correlations.} The sign reversal under partial correlation (from $0.738$ unconditional to $-0.487$ conditioning on ACL) most plausibly reflects collinearity: $\FDC$ and ACL are themselves correlated across our seven environments, since both increase with landscape smoothness, so with seven points conditioning on one absorbs most of the shared variance and leaves an unstable, sign-flipped residual for the other. We treat the partial correlations as evidence that $\FDC$ and ACL are not separable at this sample size, read the unconditional correlation as a preliminary association with landscape smoothness broadly construed, and draw no conclusion from the sign of the partial effects. We report Pearson correlation for comparability with the bootstrap CI; the Spearman correlation on the same seven environments is $\rho{=}0.571$, below the two-tailed significance threshold at $n{=}7$.

\begin{table}[h]
\centering
\caption{$\FDC$ estimator robustness on CartPole across sample sizes (10 repetitions per $N$, 95\% bootstrap CI), and the partial-correlation summary (right).}
\label{tab:fdc_robust}
\small
\begin{minipage}[t]{0.56\linewidth}
\centering
\begin{tabular}{lrrrr}
\toprule
$N$ & $\FDC$ mean & $\FDC$ std & 95\% CI & CI excl.\ 0?\\
\midrule
50 & $0.504$ & $0.135$ & $[0.242, 0.659]$ & yes\\
100 & $0.431$ & $0.093$ & $[0.266, 0.562]$ & yes\\
200 & $0.381$ & $0.084$ & $[0.274, 0.552]$ & yes\\
300 & $0.363$ & $0.082$ & $[0.260, 0.533]$ & yes\\
500 & $0.334$ & $0.093$ & $[0.245, 0.519]$ & yes\\
1000 & $0.299$ & $0.084$ & $[0.204, 0.461]$ & yes\\
\bottomrule
\end{tabular}
\end{minipage}\hfill
\begin{minipage}[t]{0.4\linewidth}
\centering
\begin{tabular}{lrr}
\toprule
Quantity & Value & $n$\\
\midrule
$r(\HD,\FDC)$, uncond.\ & $0.738$ & $7$\\
partial $|\,\mathrm{ACL}$ & $-0.487$ & $7$\\
partial $|\,\mathrm{Basin}$ & $0.551$ & $7$\\
$r(\mathrm{OrdSens},\FDC)$ & $0.832$ & $5$\\
\bottomrule
\end{tabular}
\end{minipage}
\end{table}

\begin{table}[p]
\caption{Landscape descriptors for the 15 originally-profiled environments, sorted by $\FDC$. $\dagger$: basin values of $1.0$ are threshold artifacts. OrdSens is --- for environments without a JEPA checkpoint. NR, IC, and DM are computed from random walks. The eight extended environments of Appendix~\ref{app:extended} have not yet been profiled on these descriptors. Figure~\ref{fig:ruggedness} visualizes this ranking.}
\label{tab:landfull}
\small\begin{center}
\footnotesize\setlength{\tabcolsep}{3pt}
\begin{tabular}{llrrrrrrrrr}
\toprule
Environment & Domain & $\FDC\uparrow$ & ACL$\uparrow$ & Smooth$\downarrow$ & Basin$\uparrow$ & $\HD$ & OrdSens & NR & IC & DM\\
\midrule
CartPole & balance & $0.661$ & $8$ & $9.84$ & $0.6$ & $0.218$ & $0.370$ & $0.580$ & $2.257$ & $1.327$\\
InvPendulum & balance & $0.548$ & $4$ & $14.40$ & $0.0$ & $0.050$ & $0.000$ & $0.692$ & $1.517$ & $0.864$\\
Swimmer & locomotion & $0.482$ & $3$ & $12.84$ & $0.37$ & $0.044$ & $0.068$ & $0.424$ & $2.770$ & $1.220$\\
Hopper & locomotion & $0.439$ & $5$ & $32.57$ & $0.0$ & $-0.006$ & $0.000$ & $0.982$ & $0.182$ & $2.695$\\
CliffWalking & discrete & $0.424$ & $2$ & $1.44$ & $0.0$ & --- & --- & --- & --- & ---\\
Reacher & control & $0.401$ & $5$ & $13.76$ & $0.0$ & $0.074$ & $0.003$ & $0.220$ & $2.941$ & $1.808$\\
InvDblPendulum & balance & $0.382$ & $3$ & $45.37$ & $0.0$ & --- & --- & $0.472$ & $2.858$ & $1.588$\\
Walker2d & locomotion & $0.326$ & $3$ & $47.14$ & $0.0$ & $0.000$ & $0.000$ & $0.254$ & $2.899$ & $4.566$\\
FrozenLake & discrete & $0.313$ & $1$ & --- & $0.0$ & --- & --- & $0.976$ & $0.177$ & $2.038$\\
Nim & custom & $0.248$ & $1$ & $0.15$ & $0.0$ & --- & --- & $0.718$ & $1.794$ & $3.999$\\
MountainCar(C) & control & $0.153$ & $1$ & $24.51$ & $0.1$ & $0.004$ & $-0.000$ & $0.476$ & $2.130$ & $0.734$\\
Navigation & custom & $0.130$ & $1$ & $104.85$ & $0.1$ & --- & --- & $0.130$ & $2.783$ & $1.510$\\
BallLanding & custom & $0.037$ & $4$ & $9.63$ & $0.0$ & --- & --- & $0.046$ & $2.354$ & $1.159$\\
Maze & custom & $-0.095$ & $1$ & $0.00$ & $1.000^\dagger$ & --- & --- & $0.974$ & $0.284$ & $2.038$\\
MountainCar(D) & discrete & --- & $1$ & --- & $1.000^\dagger$ & --- & --- & $0.996$ & $0.043$ & $1.312$\\
\bottomrule
\end{tabular}
\end{center}
\end{table}

\begin{figure}[h]
\centering\includegraphics[width=0.92\linewidth]{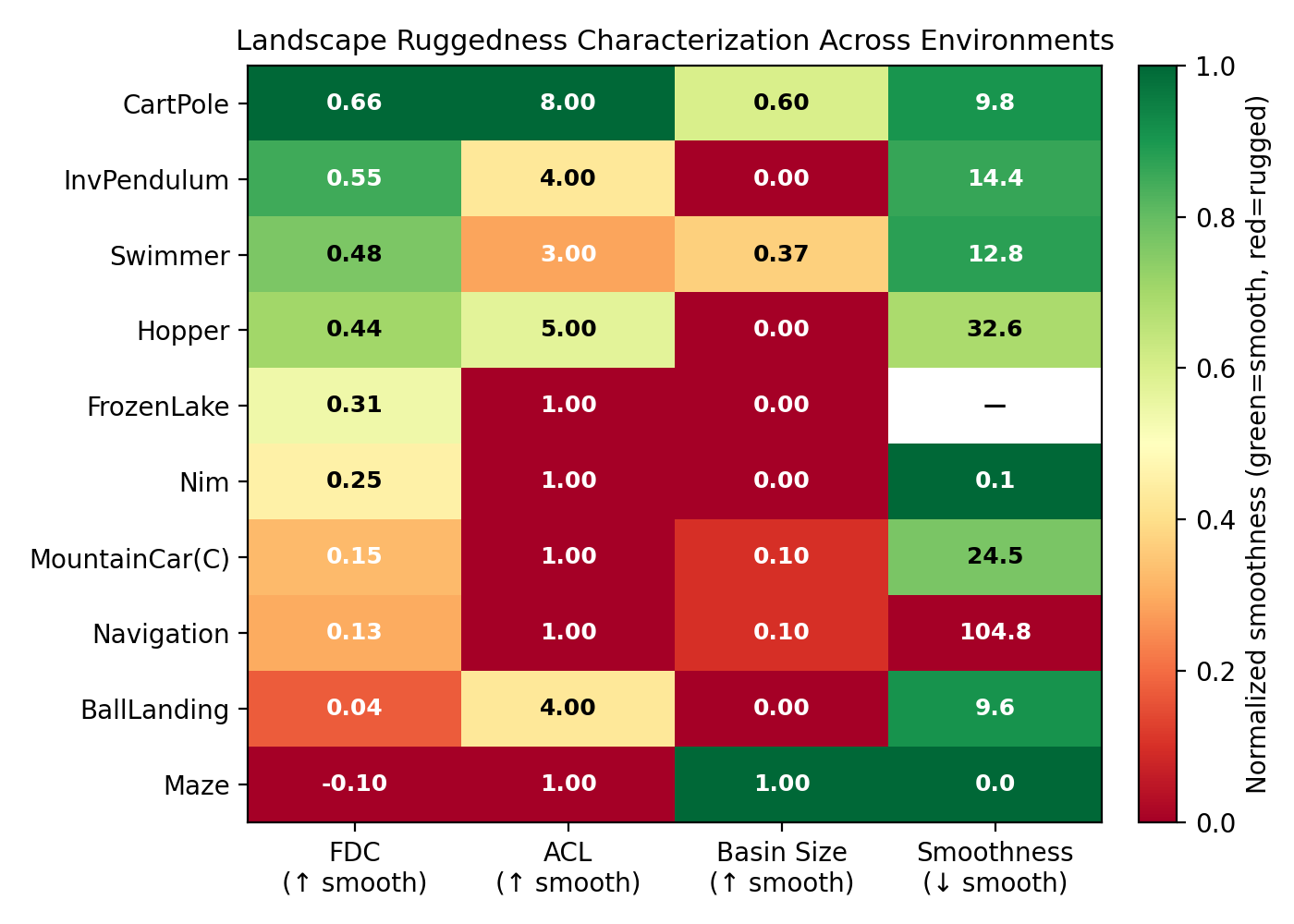}
\caption{Landscape ruggedness across the 15 originally-profiled environments (green: smoother; red: more rugged). CartPole is the most exploitable landscape; Maze, Navigation, and MountainCar the least.}
\label{fig:ruggedness}
\end{figure}

\section{History, Order, and Reward-Transformation Ablations}
\label{app:histlen}\label{app:forget}\label{app:reverse}
The history, order, reward-transformation, and inference-time perturbation ablations are consolidated in Table~\ref{tab:history_ablations}. Read the inference-time perturbations against the pretraining-time shuffle of Table~\ref{tab:shuffle} (Appendix~\ref{app:shuffle}), where the same manipulation has no effect: the two act at different stages of the pipeline. The best-only condition appears with slightly different values in three places (321.4 in Table~\ref{tab:hd}, 325.5 in the memory-condition block, 319.2 in the inference-time block), and the same reward-shuffle intervention costs 17.0\% in the inference-time block but the reward-permutation intervention costs 37\% in the reward-transformation block; these differences track the two different protocols reported alongside each table (the main $T{=}400$ protocol versus the reduced-budget, 200-iteration protocol used for the sweeps in this appendix) rather than a single consistent number, and we report each figure with the protocol it was measured under rather than reconciling them to one value.

\begin{table}[p]
\centering
\caption{History, order, and reward-transformation ablations on CartPole (and Swimmer/MountainCar where noted). The full trajectory outperforms every truncated or corrupted variant; the optimizer is sensitive to coarse temporal direction and reward rank but not to fine-grained order or reward magnitude.}
\label{tab:history_ablations}
\small

\textbf{Memory conditions vs.\ CEM} ($n{=}5$ seeds, $200$ iterations, $K{=}8$ for subset conditions)\\[2pt]
\begin{tabular}{lrrrrr}
\toprule
Condition & Full & Best-only & Recent-8 & Oldest-8 & CEM\\
\midrule
Mean best & $410.9$ & $325.5$ & $333.0$ & $317.6$ & $360.9$\\
\bottomrule
\end{tabular}

\vspace{10pt}
\textbf{Reward transformation} (mean best reward, 5 seeds; monotone transforms leave performance unchanged)\\[2pt]
\begin{tabular}{lrrr}
\toprule
Transformation & CartPole & Swimmer & MountainCar(C)\\
\midrule
Original & $337.85$ & $263.43$ & $-28.41$\\
Ranked & $338.31$ & $262.87$ & $-28.39$\\
Monotonic (log) & $337.73$ & $263.11$ & $-28.42$\\
Permuted & $212.84$ & $245.67$ & $-28.40$\\
\midrule
Ordering sensitivity (Eq.~\ref{eq:hd}) & $0.370$ & $0.068$ & $-0.000$\\
Magnitude sensitivity & $-0.001$ & $0.002$ & $-0.001$\\
\bottomrule
\end{tabular}

\vspace{10pt}
\begin{minipage}[t]{0.48\linewidth}
\centering
\textbf{Forgetting sweep} (3 seeds)\\[2pt]
\begin{tabular}{lrr}
\toprule
Condition & Mean best & $\Delta$\\
\midrule
Full history & $410.87$ & ---\\
Remove best & $285.10$ & $-30.6\%$\\
Remove early half & $300.90$ & $-26.8\%$\\
Best-only & $321.43$ & $-21.8\%$\\
Recent-8 & $309.23$ & $-24.7\%$\\
Random-8 & $320.94$ & $-21.9\%$\\
\bottomrule
\end{tabular}
\end{minipage}\hfill
\begin{minipage}[t]{0.48\linewidth}
\centering
\textbf{Order sweep} (3 seeds)\\[2pt]
\begin{tabular}{lrr}
\toprule
Condition & Mean best & $\Delta$\\
\midrule
Forward (control) & $410.87$ & ---\\
Reversed & $355.40$ & $-13.5\%$\\
Shuffled & $410.90$ & $+0.0\%$\\
\bottomrule
\end{tabular}
\end{minipage}

\vspace{10pt}
\textbf{Context-length sweep} (fixed low budget)\\[2pt]
\begin{tabular}{lrrrrrr}
\toprule
$L$ & 4 & 8 & 16 & 32 & 64 & 400\\
\midrule
Mean best & $54.5$ & $54.5$ & $54.5$ & $54.5$ & $54.5$ & $54.5$\\
\bottomrule
\end{tabular}

\vspace{10pt}
\textbf{Inference-time history perturbations} ($n{=}5$ seeds, $200$ iterations; control is the unperturbed full history)\\[2pt]
\begin{tabular}{lrrr}
\toprule
Condition & Mean best & Std & Drop vs.\ control\\
\midrule
Control (full history) & $410.9$ & $178.2$ & ---\\
Shuffle reward values & $341.1$ & $197.0$ & $-17.0\%$\\
Permute parameter dimensions & $329.7$ & $188.9$ & $-19.8\%$\\
Reverse temporal order & $354.7$ & $185.4$ & $-13.6\%$\\
Best-only (single point) & $319.2$ & $157.6$ & $-22.3\%$\\
\bottomrule
\end{tabular}
\end{table}

\section{Reward-Agnostic Pretraining and the Target Return}
\label{app:shuffle}\label{app:target}\label{app:rstar}
Table~\ref{tab:shuffle} shows that shuffling rewards, parameters, or both during pretraining leaves downstream performance unchanged, consistent with the latent-prediction objective of Eq.~\ref{eq:loss} carrying no reward-prediction term.

\begin{table}[h]
\centering
\caption{Pretraining reward/parameter shuffle (CartPole, 5 seeds).}
\label{tab:shuffle}
\small
\begin{tabular}{lrr}
\toprule
Condition & $2$k/$50$ep & $20$k/$250$ep\\
\midrule
Original (control) & $214.03$ & $207.30$\\
Rewards shuffled & $236.30$ & $208.07$\\
Params shuffled & $210.27$ & $208.00$\\
Both shuffled & $219.13$ & $207.63$\\
\bottomrule
\end{tabular}
\end{table}

Table~\ref{tab:rstar_confirmed} isolates the two reward channels at test time and shows that the target-return percentile $\rho$ has no measurable effect over the range 50 to 99 at either of two budgets. Table~\ref{tab:rstar_landscape} repeats the corruption on Swimmer and quantifies the shift in the proposal distribution. The identical means and standard deviations across conditions in Table~\ref{tab:rstar_confirmed} (410.9, 178.2) coincide with the single-agent baseline to three significant figures, which is consistent with $r^\ast$ genuinely not affecting the proposal on this landscape, but is also the pattern we would see if the reported best-over-history outcome were unaffected because it is already fixed during the random warm-up phase; the two explanations are not distinguished by this table alone.

\begin{table}[h]
\centering
\caption{Isolating the two reward channels at test time (CartPole, $n{=}5$ seeds, $T{=}400$; top), and the target-return percentile sweep at a small-scale (2k/50ep) protocol and the main protocol (bottom). Corrupting $r^\ast$ has no measurable effect at any percentile from 50 to 99, at either budget; corrupting the reward values inside the history does.}
\label{tab:rstar_confirmed}
\small
\begin{tabular}{lrrl}
\toprule
Condition & Mean & Std & $\Delta$ vs.\ normal\\
\midrule
Normal ($r^\ast{=}$ 80th percentile) & $410.9$ & $178.2$ & ---\\
$r^\ast{=}0$ & $410.9$ & $178.2$ & $0.0\%$\\
$r^\ast{=}$ random in range & $410.9$ & $178.2$ & $0.0\%$\\
$r^\ast{=}\min(r)$ (worst) & $410.9$ & $178.2$ & $0.0\%$\\
History reward values zeroed & $268.0$ & --- & $-34.8\%$\\
\bottomrule
\end{tabular}

\vspace{8pt}
\begin{tabular}{lrrrrrrr}
\toprule
$\rho$ & 50 & 60 & 70 & 80 & 90 & 95 & 99\\
\midrule
CartPole (small-scale) & $338.9$ & $339.1$ & $338.7$ & $338.5$ & $338.6$ & $338.9$ & $339.2$\\
Swimmer (small-scale)  & $178.5$ & $178.6$ & $178.7$ & $178.5$ & $178.6$ & $178.7$ & $178.7$\\
\midrule
CartPole mean (main protocol) & $410.9$ & $410.9$ & $410.9$ & $410.9$ & $410.9$ & $410.9$ & $410.9$\\
CartPole std (main protocol) & $178.2$ & $178.2$ & $178.2$ & $178.2$ & $178.2$ & $178.2$ & $178.2$\\
\bottomrule
\end{tabular}
\end{table}

\begin{table}[h]
\centering
\caption{Effect of $r^\ast$ conditioning ($n{=}5$ seeds, $T{=}400$). Disabling $r^\ast$ has no measurable effect on final reward on either landscape (top); yet varying $r^\ast$ measurably shifts the proposal distribution, as quantified by maximum mean discrepancy (MMD, bottom).}
\label{tab:rstar_landscape}
\small
\begin{tabular}{lrr}
\toprule
$r^\ast$ condition & CartPole ($\FDC{=}0.661$) & Swimmer ($\FDC{=}0.482$) \\
\midrule
Normal (80th/99th pct.) & $410.9$ & $272.1$ \\
Zero / worst / 20th pct.  & $408.5$--$412.7$ & $268.4$--$276.8$ \\
$\Delta$ vs.\ normal        & $\approx 0\%$ & $\approx 0\%$ \\
\bottomrule
\end{tabular}

\vspace{8pt}
\begin{tabular}{lr}
\toprule
$r^\ast$ percentile pair (Swimmer, $n{=}5$ seeds) & MMD \\
\midrule
20\% vs.\ 50\% & $1.762$ \\
20\% vs.\ 99\% & $1.829$ \\
80\% vs.\ 99\% & $1.633$ \\
\bottomrule
\end{tabular}
\end{table}

\section{Decoder Gradient Flow, Null Model, and Random Decoders}
\label{app:decoder}
The decoder receives exactly zero gradient, since $\mathcal{L}_{\mathrm{JEPA}}$ compares two latent vectors and never routes gradient through $h_\xi$, which is invoked only inside the no-gradient path used at inference (Table~\ref{tab:decoder_ablation}, bottom). We freeze the encoder and predictor and fine-tune $h_\xi$ for 50 epochs with $\mathcal{L}_{\mathrm{recon}}=\|h_\xi(\hat s_y)-\theta_{t+1}\|_2^2$; the reconstruction loss decreases as expected, but downstream performance falls from 410.9 to 293.7. Both trained variants underperform the frozen decoder by a substantial margin, which is inconsistent with a mean-plus-noise account of JEPA-PS (a decoder trained to approximate the relevant statistic should do at least as well as an untrained one) and consistent with the encoder and predictor arranging a latent geometry in which a fixed random projection already lands near useful parameters.

\begin{table}[h]
\caption{Effect of decoder training on downstream reward (CartPole, $n{=}5$ seeds), and mean gradient norm reaching each component during a training step.}
\label{tab:decoder_ablation}
\small\begin{center}
\begin{tabular}{lrr}
\toprule
Decoder variant & Reward & $\Delta$ vs.\ frozen \\
\midrule
Frozen random (default) & $410.9$ & --- \\
Reconstruction fine-tuned & $293.7$ & $-28.5\%$ \\
End-to-end joint retraining & $269.0$ & $-34.5\%$ \\
\bottomrule
\end{tabular}

\vspace{6pt}
\begin{tabular}{lcc}
\toprule
Component & Mean gradient norm & Role\\
\midrule
Context encoder $g_\phi$ & $0.2479$ & encodes the history\\
Predictor $p_\psi$ & $0.4533$ & predicts the target latent\\
Decoder $h_\xi$ & $0.0000$ & maps the latent to $\hat\theta$\\
\bottomrule
\end{tabular}
\end{center}
\end{table}

\paragraph{Null model: two protocols, two directions of result.} Table~\ref{tab:null_model} compares JEPA-PS with a proposer sampling $\mathcal{N}(\mu_h,\sigma_h)$ from the history statistics alone under our main evaluation protocol, where JEPA-PS leads on both environments. We separately ran a matched-protocol head-to-head on Swimmer, with identical warm-starts and an identical proposal-noise seed shared across both conditions ($n{=}10$): under this stricter design, the null proposer reached $170.5{\pm}77.3$ against $133.3{\pm}73.3$ for JEPA-PS, reversing the direction of the comparison. A constant-latent control (the same frozen decoder fed a single fixed input regardless of history) scored $137.8{\pm}70.1$, statistically indistinguishable from JEPA-PS. We report both results rather than retain only the first: under our main protocol the latent appears to add value beyond history statistics; under the more tightly matched protocol it does not, on Swimmer specifically. 

\begin{table}[h]
\caption{JEPA-PS vs.\ a null statistics-only proposer under two protocols. Top: main protocol ($n{=}10$ seeds, $T{=}400$). Bottom: matched-protocol head-to-head with shared warm-starts and proposal-noise seed ($n{=}10$, Swimmer only), which reverses the direction of the comparison.}
\label{tab:null_model}
\small\begin{center}
\begin{tabular}{lrr}
\toprule
Method (main protocol) & CartPole & Swimmer \\
\midrule
Null $\mathcal{N}(\mu_h,\sigma_h)$ & $302.1$ & $151.9$ \\
JEPA-PS & $410.9$ & $293.1$ \\
\bottomrule
\end{tabular}

\vspace{6pt}
\begin{tabular}{lr}
\toprule
Method (matched protocol, Swimmer only) & Mean $\pm$ std \\
\midrule
Null $\mathcal{N}(\mu_h,0.3\sigma_h)$ & $\mathbf{170.5{\pm}77.3}$ \\
Constant-latent control & $137.8{\pm}70.1$ \\
JEPA-PS & $133.3{\pm}73.3$ \\
\bottomrule
\end{tabular}
\end{center}
\end{table}

\paragraph{Ten independent random decoders.} We draw ten independent random linear decoders and run Latent-CEM. Nine of the ten draws give $410.9{\pm}178.2$, identical to the default decoder to three significant figures; one draw reaches the environment ceiling. Across all ten the mean is 419.8 with a between-decoder standard deviation of 26.7, small relative to the within-decoder seed variance of 178.2, so performance under this ablation is governed by the encoder--predictor latent geometry rather than by the specific decoder draw; we note this finding concerns the decoder's identity specifically and does not resolve the matched-protocol null-model reversal above, which concerns the latent's content.

\section{Population Wrapper Ablations}
\label{app:pop}\label{app:budgetfair}
\begin{figure}[h]
\centering\includegraphics[width=0.92\linewidth]{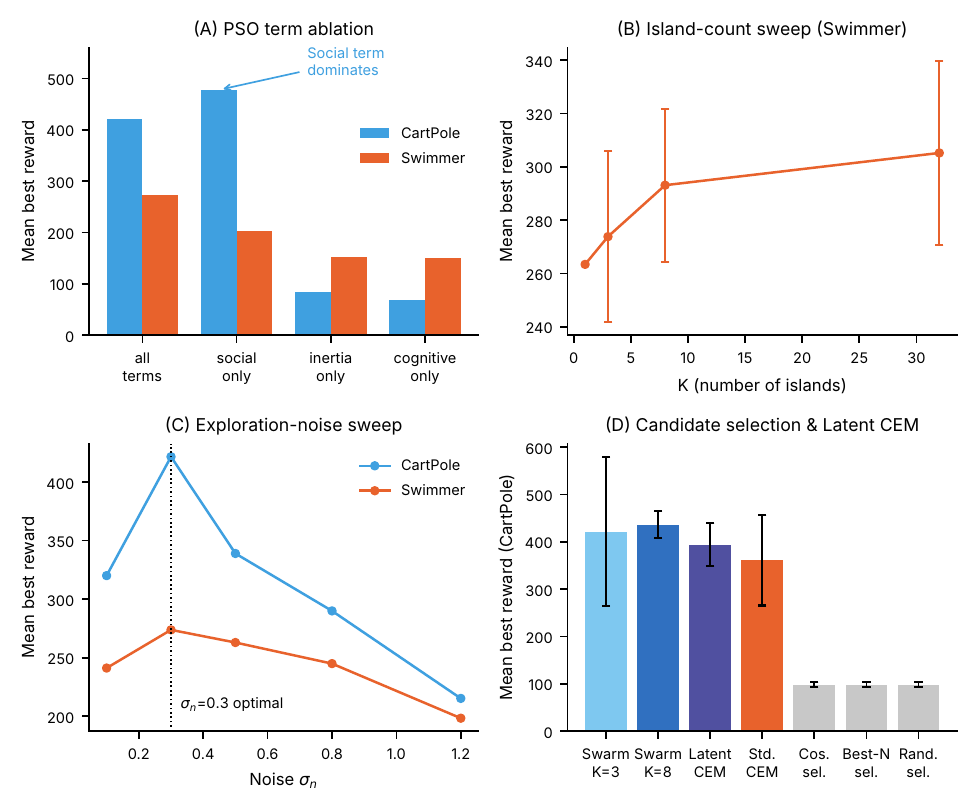}
\caption{Population-wrapper ablations. \textbf{(A)} Swarm-term ablation on CartPole and Swimmer. \textbf{(B)} Island-count sweep on Swimmer: the mean rises and variance falls, then widens, with diminishing returns. \textbf{(C)} Exploration-noise sweep, with $\sigma_n{=}0.3$ best on both tasks. \textbf{(D)} Candidate-selection rule and Latent-CEM comparison on CartPole.}
\label{fig:ablations}
\end{figure}

Figure~\ref{fig:ablations} ablates the three swarm terms, sweeps the island count, and sweeps the exploration noise; Table~\ref{tab:budget_fairness} lists Swarm-JEPA's sensitivity to migration frequency and exploration noise, and compares fixed-total against fixed-per-island budget accounting. On the easy task the social term alone suffices; on the harder task all three contribute. Island count improves the mean and reduces variance with diminishing returns, and $\sigma_n=0.3$ is best on both tasks. Separately, we found that this migration mechanism does not supply genuine cross-island diversity at any population size we swept (Appendix~\ref{app:invpend_diag}), which bears on InvertedPendulum's failure specifically. Note that this population wrapper is architecturally distinct from the restart-on-stagnation wrapper of Section~\ref{sec:transfer}: Swarm-JEPA runs $K$ agents in parallel with periodic migration, while restart-on-stagnation reinitializes a single agent's history after a fixed number of non-improving steps; the two are not interchangeable and we did not test Swarm-JEPA under the Section~\ref{sec:transfer} attribution protocol.

\paragraph{Budget accounting.} At $K{=}3$ the fixed-total and fixed-per-island schemes give identical results (421.6), so the gain of $K{=}3$ over $K{=}1$ (410.9) is attributable to population structure rather than to additional evaluations. At $K{=}8$, fixed-total accounting reproduces the headline result of Table~\ref{tab:main} (416.1), while fixed-per-island accounting, which grants eight times the actual evaluation budget, reaches the environment ceiling; the gap is therefore attributable to the extra budget itself.

\begin{table}[h]
\centering
\caption{Swarm-JEPA budget accounting and hyperparameter sensitivity (CartPole). Top: fixed-total vs.\ fixed-per-island budget accounting ($n{=}5$ seeds). Bottom: sensitivity to migration frequency and exploration noise ($K{=}3$, $n{=}3$ seeds, $400$ iterations).}
\label{tab:budget_fairness}
\small
\begin{tabular}{lccccc}
\toprule
$K$ & Mode & Iters/island & Total evals & Mean best & Std\\
\midrule
1 & either & $400$ & $400$ & $410.9$ & $178.2$\\
3 & fixed-total & $133$ & $400$ & $421.6$ & $156.9$\\
3 & fixed-per-island & $400$ & $1200$ & $421.6$ & $156.9$\\
8 & fixed-total & $50$ & $400$ & $416.1$ & $28.7$\\
8 & fixed-per-island & $400$ & $3200$ & $500.0$ & $0.0$\\
\bottomrule
\end{tabular}

\vspace{10pt}
\begin{tabular}{lrr}
\toprule
Configuration & Mean best & Std\\
\midrule
\multicolumn{3}{l}{\emph{Migration frequency}}\\
Every 10 steps & $413.1$ & $123.6$\\
Every 20 steps & $417.8$ & $150.3$\\
Every 30 steps (default) & $421.6$ & $156.9$\\
Every 60 steps & $421.6$ & $156.9$\\
No migration & $421.6$ & $156.9$\\
Single agent ($K{=}1$) & $410.9$ & $178.2$\\
\midrule
\multicolumn{3}{l}{\emph{Exploration noise $\sigma_n$}}\\
0.1 & $423.2$ & $163.7$\\
0.3 (default) & $421.6$ & $156.9$\\
0.5 & $420.8$ & $139.3$\\
1.0 & $399.1$ & $152.0$\\
\bottomrule
\end{tabular}
\end{table}

\section{Robustness and Sample Efficiency}
\label{sec:robustness}\label{app:robustfull}
We stress-test Swarm-JEPA under restricted budgets, larger populations, reduced pretraining data, and corrupted reward signals. Unless stated otherwise, experiments use CartPole; Swimmer and Hopper replication follows (Tables~\ref{tab:robustness_swimmer}--\ref{tab:robustness_hopper}).

\paragraph{A note on CartPole variance.} CartPole's mean best reward varies across tables (roughly 337, 410.9, 416.1, 421.6, 459.0, and 480--492). Budget and protocol differences account for the largest swings: 337 under a reduced budget (200 iterations, 5 episodes per evaluation, Table~\ref{tab:history_ablations}), 410.9 (single-agent) under the main $T{=}400$ protocol, and 480--492 in the 10-seed, high-trajectory-count data-efficiency experiment (Table~\ref{tab:robustness_cartpole}). Under the identical $T{=}400$ protocol, 410.9 is single-agent JEPA-PS, 421.6 is Swarm-JEPA ($K{=}3$), and 416.1 is Swarm-JEPA ($K{=}8$); these are different population sizes sharing a fixed total budget. The environment's difficulty is bimodal, with most seeds reaching the ceiling of 500 and a minority failing to, so standard deviations of 150--200 are a property of the task rather than of any one method. We separately found, using an exact distribution-free test based on record statistics, that on several environments the probability a run's reported best is determined entirely by its random warm-up phase (i.e.\ the optimizer never subsequently improves on it) is far higher than chance would predict if post-warmup proposals were simply as good as continued random search ($p<10^{-4}$ on every environment tested); this does not change the numbers reported here but means comparisons on best-over-history specifically should be interpreted with this floor in mind.

\paragraph{Budget scaling.} Swarm-JEPA rises from 283.8 at ten evaluations to 404.9 at fifty and saturates near 410.9 from one hundred onward. CEM matches JEPA's 100-evaluation performance at that budget and reaches 477.1 by the full budget: JEPA-PS converges faster early, but CEM catches up sooner and plateaus higher on this landscape. We additionally tested, on a timestep-matched locomotion suite (HalfCheetah, Swimmer, Hopper, Ant, Walker2d), whether JEPA-PS's pretrained prior confers a low-budget sample-efficiency advantage over ARS, CEM, and a CMA-ES-like baseline; it does not, trailing all three from the smallest budget tested onward, so we do not claim a sample-efficiency advantage.

\begin{figure}[h]
\centering\includegraphics[width=0.92\linewidth]{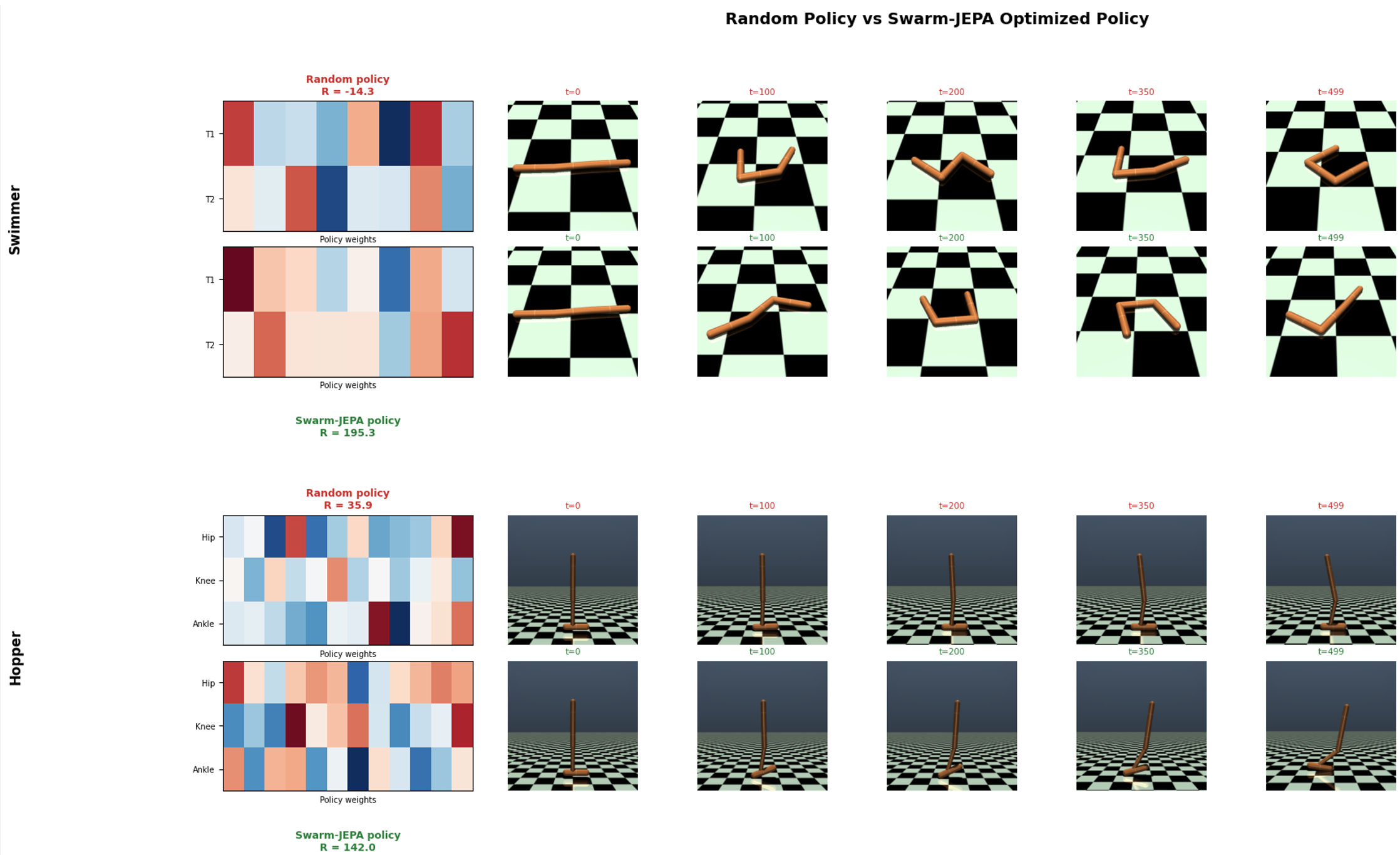}
\caption{Policy visualization on Swimmer and Hopper. Weight matrices (left) evolve from unstructured initialization to action-specific structure; rendered frames (right) show behavior across an episode. A random Swimmer policy flails without net forward motion while an optimized policy produces coordinated undulation; a random Hopper policy collapses immediately while an optimized policy sustains hopping.}
\label{fig:robot_renders}
\end{figure}

\paragraph{Population size.} On Swimmer under a fixed total budget, performance increases monotonically from 263.4 at one island to 305.2 at 32 islands. The standard deviation contracts from 28.71 at eight islands to 19.30 at sixteen and widens again to 34.60 at thirty-two. Figure~\ref{fig:robot_renders} visualizes the resulting Swimmer and Hopper policies qualitatively.

\paragraph{Data efficiency.} Every checkpoint trained on 300 or more trajectories reaches at least 481, approaching the full 20,000-trajectory model (492.3); even 100 trajectories yield 480.2. This near-saturation reflects CartPole's smoothness; data efficiency on a rougher landscape remains open.

\paragraph{Noisy, sparse, and delayed rewards.} On CartPole, degradation under Gaussian reward noise is non-monotone (427.2 at 50\% noise vs.\ 270.0 at 5\%), within one standard deviation and not significant, so we do not interpret the ordering; the robust observation is the absence of monotone collapse. Feedback only every $k$ evaluations costs 35\% between $k{=}1$ and $k{=}5$ and then plateaus. A short delay costs 21\%, while longer delays recover and exceed the undelayed baseline (487.8 at 50 steps). The replication on Swimmer and Hopper is less benign: on Hopper, 10\% noise, 50\% noise, and sparse feedback all yield $127.6{\pm}69.9$, close to the single-agent result (133--136), and on Swimmer 10\% noise costs about half the baseline. Robustness to corrupted rewards is therefore environment-dependent.

\paragraph{Non-stationary objectives.} When the reward transformation is switched partway through a run (sign flip, rescaling, or additive offset), performance collapses to the random-policy floor: the history encodes the original objective. Detecting distribution shift and warm-restarting is a natural extension; Appendix~\ref{app:cmaes_battery}'s Exp.\ 3/3b show that, where CMA-ES can adapt online to such a shift, JEPA-PS's recovery is far slower and less complete.

\begin{table}[p]
\centering
\caption{Full CartPole robustness grid: evaluation-budget scaling, Gaussian reward noise, pretraining data efficiency, sparse feedback, and delayed feedback ($n{=}5$ seeds unless noted).}
\label{tab:robustness_cartpole}
\small

\textbf{Budget scaling} (Swarm-JEPA saturates near 100 evaluations; CEM overtakes it by 200)\\[2pt]
\begin{tabular}{lrrrrrr}
\toprule
Method & 10 & 20 & 50 & 100 & 200 & 400\\
\midrule
Swarm-JEPA & $283.8$ & $328.1$ & $404.9$ & $410.9$ & $410.9$ & $410.9$\\
CEM & $224.1$ & $278.3$ & $278.3$ & $413.9$ & $447.6$ & $483.1$ \\
\bottomrule
\end{tabular}

\vspace{10pt}
\textbf{Gaussian reward noise} (reward range $\approx410$; degradation is non-monotone and not statistically significant)\\[2pt]
\begin{tabular}{lrrrrr}
\toprule
Noise level & 0\% & 5\% & 10\% & 20\% & 50\%\\
\midrule
Mean best & $410.9$ & $270.0$ & $239.3$ & $287.8$ & $427.2$\\
\bottomrule
\end{tabular}

\vspace{10pt}
\textbf{Pretraining data efficiency} (both mean reward and variance improve smoothly with corpus size)\\[2pt]
\begin{tabular}{lrrrrrrr}
\toprule
$n$ trajectories & 100 & 300 & 500 & 1{,}000 & 3{,}000 & 5{,}000 & 20{,}000\\
\midrule
Mean best & $480.2$ & $481.1$ & $483.3$ & $487.6$ & $488.1$ & $490.1$ & $492.3$\\
Std & $29.6$ & $43.5$ & $27.6$ & $24.5$ & $22.1$ & $18.5$ & $15.4$\\
\bottomrule
\end{tabular}

\vspace{10pt}
\begin{minipage}[t]{0.46\linewidth}
\centering
\textbf{Sparse feedback}\\[2pt]
\setlength{\tabcolsep}{3pt}\begin{tabular}{lrrrr}
\toprule
Feedback every $k$ steps & 1 & 5 & 10 & 20\\
\midrule
Mean best & $410.9$ & $268.0$ & $268.0$ & $268.0$\\
\bottomrule
\end{tabular}
\end{minipage}\hfill
\begin{minipage}[t]{0.5\linewidth}
\centering
\textbf{Delayed feedback} (a short delay costs performance; longer delays recover and exceed it)\\[2pt]
\setlength{\tabcolsep}{3pt}\begin{tabular}{lrrrrr}
\toprule
Delay $d$ (steps) & 0 & 5 & 10 & 20 & 50\\
\midrule
Mean best & $410.9$ & $324.1$ & $452.9$ & $435.2$ & $487.8$\\
\bottomrule
\end{tabular}
\end{minipage}
\end{table}

\begin{table}[h]
\begin{minipage}[t]{0.48\linewidth}
\centering
\caption{Robustness on Swimmer-v4 ($n{=}5$ seeds, $T{=}400$). A 50-step delay substantially improves performance relative to the short-delay and sparse conditions, though it remains slightly below the no-perturbation baseline.}
\label{tab:robustness_swimmer}
\small
\begin{tabular}{lrr}
\toprule
Condition & Mean & Std \\
\midrule
Baseline (noise 0\%)  & $293.1$ & $28.7$ \\
Noise 10\%            & $150.6$ & $25.7$ \\
Noise 20\%            & $156.2$ & $18.5$ \\
Noise 50\%            & $190.6$ & $25.4$ \\
Sparse $k{=}5$        & $93.9$  & $62.0$ \\
Sparse $k{=}10$       & $93.9$  & $62.0$ \\
Sparse $k{=}20$       & $97.2$  & $67.0$ \\
Delay $d{=}5$         & $96.8$  & $59.4$ \\
Delay $d{=}10$        & $100.4$ & $56.0$ \\
Delay $d{=}20$        & $103.0$ & $56.0$ \\
Delay $d{=}50$        & $274.9$ & $89.3$ \\
\bottomrule
\end{tabular}
\end{minipage}\hfill
\begin{minipage}[t]{0.48\linewidth}
\centering
\caption{Robustness on Hopper-v4 ($n{=}5$ seeds, $T{=}400$).}
\label{tab:robustness_hopper}
\small
\begin{tabular}{lrr}
\toprule
Condition & Mean & Std \\
\midrule
Baseline (noise 0\%)  & $1039.7$ & $33.5$ \\
Noise 10\%            & $127.6$ & $69.9$ \\
Noise 50\%            & $127.6$ & $69.9$ \\
Sparse $k{=}5$        & $127.6$ & $69.9$ \\
Delay $d{=}50$        & $345.3$ & $328.9$ \\
\bottomrule
\end{tabular}
\end{minipage}
\end{table}

\section{LLM Optimizer Comparison}
\label{app:props}
\begin{figure}[h]
\centering\includegraphics[width=0.92\linewidth]{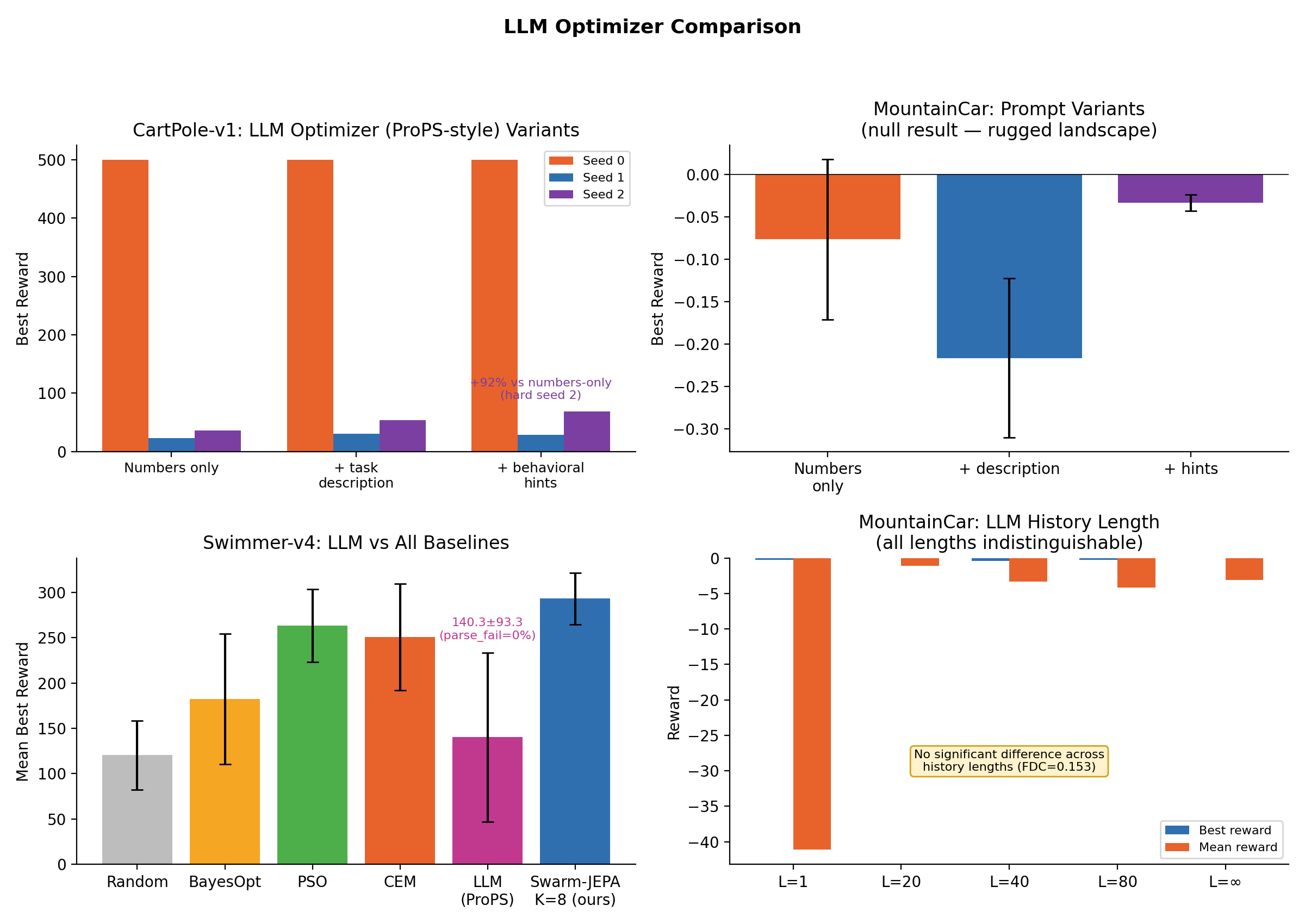}
\caption{LLM optimizer comparison. \textbf{(A)} CartPole across three seeds: hints help on the hard seed and not on the easy one. \textbf{(B)} MountainCar: all prompt variants remain near the random floor. \textbf{(C)} Swimmer: the LLM optimizer trails the population wrapper. \textbf{(D)} History-length sweep on MountainCar.}
\label{fig:props}
\end{figure}

The LLM optimizer solves the easy CartPole seed under every prompt configuration and separates only on the harder seeds, where behavioral hints help (Table~\ref{tab:props_cartpole}; summarized visually in Figure~\ref{fig:props}). On MountainCar every configuration remains near the random floor regardless of prompt content or history length (Table~\ref{tab:props_mc}), consistent with that task's low FDC. On Swimmer it trails the population wrapper at 77 times higher per-step cost. On Hopper, the two are comparable (Table~\ref{tab:main}). Section~\ref{sec:llm} gives the main-text treatment of this comparison, including two further, independently run LLM-agent baselines and a restart-wrapper control.

\begin{table}[h]
\caption{LLM optimizer vs.\ Swarm-JEPA on CartPole and Swimmer.}
\label{tab:props_cartpole}
\centering\small
\begin{minipage}[t]{0.48\linewidth}
\centering
\textbf{CartPole (best reward, 3 seeds)}\\[2pt]
\begin{tabular}{lrrr}
\toprule
Mode & Seed 0 & Seed 1 & Seed 2\\
\midrule
Numbers only & $500$ & $23$ & $36$\\
$+$ task description & $500$ & $30$ & $54$\\
$+$ behavioral hints & $500$ & $29$ & $69$\\
Swarm-JEPA ($K{=}8$) & $500$ & $500$ & $248$\\
\bottomrule
\end{tabular}
\end{minipage}\hfill
\begin{minipage}[t]{0.48\linewidth}
\centering
\textbf{Swimmer (3 seeds)}\\[2pt]
\setlength{\tabcolsep}{3pt}\begin{tabular}{lrrrr}
\toprule
Method & S0 & S1 & S2 & Mean$\pm$Std\\
\midrule
LLM (Qwen2.5-7B) & $240.1$ & $124.7$ & $56.0$ & $140.3{\pm}93.3$\\
Swarm-JEPA ($K{=}8$) & --- & --- & --- & $293.1{\pm}28.7$\\
\bottomrule
\end{tabular}
\end{minipage}
\end{table}

\begin{table}[h]
\caption{MountainCar: LLM optimizer robustness across seeds, prompt variants, and history length. All configurations remain near the random floor ($\FDC{=}0.153$).}
\label{tab:props_mc}
\centering\small
\begin{minipage}[t]{0.48\linewidth}
\centering
\textbf{By seed}\\[2pt]
\setlength{\tabcolsep}{3pt}\begin{tabular}{lrrrr}
\toprule
Mode & S0 & S1 & S2 & Mean\\
\midrule
Numbers only & $-0.21$ & $-0.02$ & $-0.00$ & $-0.08$\\
$+$ description & $-0.10$ & $-0.22$ & $-0.33$ & $-0.22$\\
$+$ hints & $-0.04$ & $-0.04$ & $-0.02$ & $-0.03$\\
Swarm-JEPA ($K{=}3$) & $-1.5$ & $-6.2$ & $-6.9$ & $-4.9$\\
\bottomrule
\end{tabular}
\end{minipage}\hfill
\begin{minipage}[t]{0.48\linewidth}
\centering
\textbf{By prompt variant (seed 0)}\\[2pt]
\begin{tabular}{lrrrr}
\toprule
Mode & Orig. & Var.\,1 & Var.\,2 & Var.\,3\\
\midrule
Numbers only & $-0.09$ & $-0.15$ & $-0.23$ & $-0.40$\\
$+$ description & $-0.40$ & $-0.07$ & $-0.16$ & $-0.40$\\
\bottomrule
\end{tabular}
\end{minipage}

\vspace{8pt}
\centering
\textbf{By history length (seed 0)}\\[2pt]
\begin{tabular}{lrrrrr}
\toprule
Max history & $L{=}1$ & $L{=}20$ & $L{=}40$ & $L{=}80$ & Unbounded\\
\midrule
Best reward & $-0.28$ & $-0.03$ & $-0.40$ & $-0.25$ & $-0.04$\\
Mean reward & $-41.12$ & $-1.12$ & $-3.30$ & $-4.14$ & $-3.10$\\
\bottomrule
\end{tabular}
\end{table}

\section{Latent-Space Search and Candidate Selection}
\label{app:latentcem}
Running CEM's elite-resampling update inside the learned latent space rather than in raw parameter space improves mean best reward by 9.1\% on CartPole and 41.6\% on Swimmer (Table~\ref{tab:latentcem}), indicating that the latent space is a more tractable search manifold than the raw parameters. Replacing cosine-similarity candidate selection with best-of-$n$ or random selection changes nothing on CartPole, so the explicit selection rule contributes little beyond the decoder's proposal.

\begin{table}[h]
\caption{Latent-space CEM vs.\ raw-space CEM (mean best reward, 3 seeds).}
\label{tab:latentcem}
\small\begin{center}
\begin{tabular}{lrr}
\toprule
Method & CartPole & Swimmer\\
\midrule
CEM (raw parameter space) & $360.86{\pm}95.3$ & $215.00{\pm}42.1$\\
Latent-CEM (JEPA latent) & $393.71{\pm}45.2$ & $304.48{\pm}38.5$\\
\bottomrule
\end{tabular}
\end{center}
\end{table}

\section{Standardized Cross-Method Comparison}
\label{app:standardized}
To provide a single point of comparison unaffected by the protocol differences discussed in Appendix~\ref{sec:robustness}, Table~\ref{tab:standardized} fixes $T{=}400$ evaluations, $n{=}10$ seeds, and $10$ episodes per evaluation for every method on CartPole. CEM, CMA-ES, and PSO reuse the tuned $n{=}10$ figures from Table~\ref{tab:strong_grid}; Random Search has no separate tuned variant, so its untuned $n{=}5$ result from Table~\ref{tab:main} is reported. The median of 500.0 against a JEPA mean of 410.9 reflects the bimodal difficulty of the environment: most seeds solve it outright and a minority do not.

\begin{table}[h]
\centering
\caption{Standardized CartPole comparison: $T{=}400$, $n{=}10$ seeds, $10$ episodes per evaluation.}
\label{tab:standardized}
\small
\begin{tabular}{lrrr}
\toprule
Method & Mean & Std & Median\\
\midrule
Swarm-JEPA (single agent) & $410.9$ & $178.2$ & $500.0$ \\
CEM & $484.2$ & $31.7$ & $500.0$ \\
CMA-ES & $500.0$ & $0.0$ & $500.0$ \\
PSO & $500.0$ & $0.0$ & $500.0$ \\
Random & $500.0$ & $0.0$ & $500.0$ \\
\bottomrule
\end{tabular}
\end{table}

\section{Additional Mechanistic Observations}
\label{app:mech}
\begin{figure}[h]
\centering\includegraphics[width=0.92\linewidth]{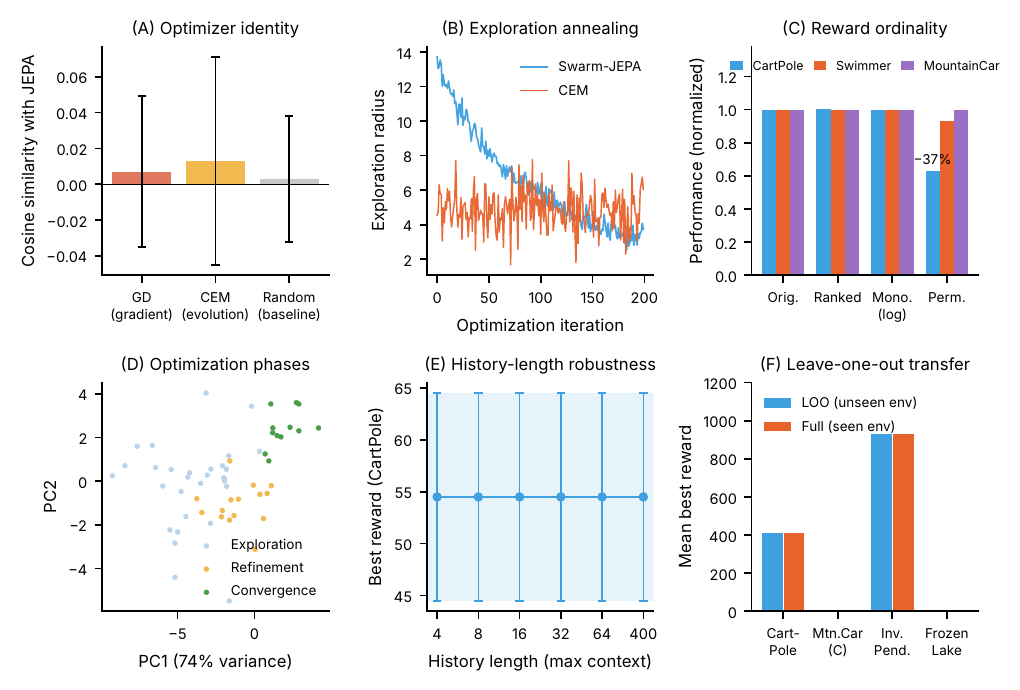}
\caption{Secondary mechanistic observations. \textbf{(A)} Cosine similarity between JEPA proposals and classical update directions, all near zero. \textbf{(B)} Exploration radius contracting over a run. \textbf{(C)} History-length robustness. \textbf{(D)} Leave-one-out transfer.}
\label{fig:novel_findings}
\end{figure}

For completeness we record secondary observations that are consistent with, but not central to, the paper's claims. Proposal directions are close to orthogonal to finite-difference gradient ascent, CEM's elite-mean update, PSO, expected improvement, upper confidence bound, the natural gradient, and a random direction (Figure~\ref{fig:novel_findings}; all cosine similarities lie between $0.003$ and $0.015$). The exploration radius contracts over the course of a run with no programmed schedule (correlation of $-0.590$ with iteration count). The latent trajectory concentrates on a near-one-dimensional manifold and separates into clusters aligned with the early, middle, and late phases of optimization. None of these is load-bearing for Section~\ref{sec:landscape}. A separate analysis searching for signatures of chaotic dynamics in optimizer trajectories on NK landscapes (largest Lyapunov exponent, FFT power spectra) found no relationship supporting that explanation for the landscape-dependent effects reported in this paper and is not pursued further.

\section{A Non-Control Black-Box Setting: BBOB}
\label{app:bbob}
To probe generality beyond control tasks, we run the JEPA optimizer unmodified, without additional pretraining, on five standard BBOB synthetic functions ($d{=}8$) against CEM (Table~\ref{tab:bbob}). The optimizer is competitive only on the unimodal Sphere function and is clearly outperformed by CEM on Ellipsoid, Rosenbrock, and Rastrigin; the Schwefel result is unreliable owing to a domain mismatch between the function's native range and the scale seen in pretraining. BBOB functions have structural properties (Rosenbrock's curved narrow valley, Rastrigin's regular periodic multimodality) quite unlike the funnel-shaped control landscapes used for pretraining. This establishes a boundary condition: the model as pretrained here is a control-task optimizer, not a general-purpose black-box optimizer.

\paragraph{Failure severity tracks landscape smoothness, and pretraining on BBOB trajectories recovers most of the gap.} Table~\ref{tab:bbob} also reports $\FDC$ measured directly on each BBOB function: Sphere is by far the smoothest, followed by Rosenbrock, with Ellipsoid, Rastrigin, and Schwefel all rugged. This ordering tracks the severity of JEPA-PS's failure: the optimizer is competitive only on Sphere, the one function with high FDC. We then pretrain a checkpoint directly on BBOB rollouts (no control-task data), sampled across all five functions, and re-evaluate. Measured as a reduction in the deficit to CEM, the BBOB-pretrained model closes most of the gap on Ellipsoid (deficit from 115,775 to 57,215, a 51\% reduction) and on Rosenbrock (from 372.1 to 6.9, a 98\% reduction); Rastrigin is essentially unchanged and Schwefel worsens numerically, consistent with Schwefel's low FDC making it hard for any fixed-budget in-context method. The transfer failure thus reflects a mismatch between pretraining distribution and test landscape rather than an architectural ceiling -- and is itself a further instance of the same pattern as Section~\ref{sec:transfer}: pretrained structure transfers within a family of related landscapes and does not transfer across structurally different ones without re-exposure. Section~\ref{sec:transfer}, Part~7 extends this specific finding: a curriculum deliberately broadened to include NK landscapes spanning smooth to rugged, not just BBOB functions, still fails to transfer to Rastrigin ($-49.5\pm8.1$, statistically indistinguishable from both numbers in this table), so the failure is not simply fixed by adding more landscape diversity of the kind we tried.

\begin{table}[h]
\centering
\caption{JEPA optimizer vs.\ CEM on BBOB synthetic functions ($d{=}8$, $n{=}5$ seeds, $400$ evaluations; higher is better as scored), with $\FDC$ measured directly on each function ($300$ random samples) and results for a checkpoint pretrained directly on BBOB rollouts (all 5 functions mixed, no control-task data).}
\label{tab:bbob}
\footnotesize\setlength{\tabcolsep}{3pt}
\begin{tabular}{lrrrrrr}
\toprule
Function & $\FDC$ & JEPA & JEPA std & JEPA (BBOB-pretrained) & CEM mean & CEM std\\
\midrule
Sphere (f1) & $0.861$ & $-2.12$ & $0.91$ & $-1.19$ & $-1.32$ & $1.01$\\
Ellipsoid (f2) & $0.250$ & $-119190$ & $31871$ & $-60630$ & $-3415$ & $2340$\\
Rosenbrock (f8) & $0.649$ & $-524.3$ & $112.1$ & $-159.1$ & $-152.2$ & $110.2$\\
Rastrigin (f15) & $0.215$ & $-53.0$ & $10.7$ & $-53.0$ & $-16.2$ & $5.7$\\
Schwefel (f20) & $0.109$ & $-10.2$ & $5.3$ & $-1959.4$ & $2057$ & $821$\\
\bottomrule
\end{tabular}
\end{table}

\section{InvertedPendulum Diagnostic Battery}
\label{app:invpend_diag}
We applied the diagnostic battery used to investigate failure modes on other environments to the InvertedPendulum checkpoint, motivated by the gap discussed in Section~\ref{sec:results}. Across 20 seeds, JEPA-PS's proposals moved consistently toward a good-solution region estimated from independent random probing (mean distance to that region: $1.81$ post-warmup vs.\ $2.65$ during warmup), rather than away from it as in the divergence pathology characterized for other environments. The context encoder's representation diversity across 40 distinct histories and the decoder's weight-matrix conditioning were both unremarkable relative to other checkpoints. A 300-sample random probe of InvertedPendulum's parameter space locates a best reward of only around 58, some 15-fold below IPOP-CMA-ES's eventual 890.0. IPOP-CMA-ES reaches this region through its population-doubling restart schedule \citep{auger2005restart}; JEPA-PS's single-agent form has no analogous restart or basin-hopping mechanism, and the Swarm-JEPA population wrapper does not supply one either, since its migration step synchronizes islands before any could independently locate a narrow basin (confirmed separately via an island-count sweep that found performance flat across $K{=}1$ through $K{=}16$, consistent with the migration schedule equalizing islands' histories faster than diversity can accumulate). We conclude the gap reflects the interaction between IPOP-CMA-ES's restart mechanism and a narrow high-reward basin that a fixed random-probe budget does not reliably locate, rather than a defect in the proposal mechanism. This is consistent with, and anticipates, Section~\ref{sec:transfer}'s later finding that a different restart mechanism (restart-on-stagnation, introduced there) substantially improves JEPA-PS's reach on CartPole physics variants: restarts matter for this architecture generally, not only on InvertedPendulum. This concerns whether the optimizer can \emph{discover} a hard-to-reach basin, which is a question about basin discovery, not about how much of an already-discovered trajectory the optimizer exploits; it also means that $\HD$ and $\OS$ on this environment are measured on an optimizer that is not making progress, a caveat we carry forward into Section~\ref{sec:landscape}.

\end{document}